\documentclass{article}

\usepackage[preprint]{corl_2024} %

\usepackage{amsmath}
\usepackage{amssymb}
\usepackage{mathtools}
\usepackage{amsthm}

\usepackage[capitalize,noabbrev]{cleveref}

\theoremstyle{plain}

\theoremstyle{definition}

\theoremstyle{remark}

\usepackage{hyperref}
\usepackage{url}

\usepackage[utf8]{inputenc} 
\usepackage[T1]{fontenc}    

\usepackage{scalefnt,letltxmacro}
\LetLtxMacro{\oldtextsc}{\textsc}
\renewcommand{\textsc}[1]{\oldtextsc{\scalefont{1.10}#1}}

\usepackage{caption}
\usepackage{subcaption}

\usepackage{microtype}
\usepackage{textgreek}
\usepackage{amsfonts,amsmath,amssymb}   %
\usepackage{graphicx}         %
\usepackage{booktabs,tabularx}          %
\usepackage{nicefrac}                   %
\usepackage{graphics}                   %
\usepackage{multirow}
\usepackage{lipsum}  
\usepackage[font=footnotesize]{caption}
\usepackage{algorithm}
\usepackage{algorithmic}
\usepackage{adjustbox}
\usepackage{soul}
\usepackage{xcolor}
\usepackage{pdfpages}
\usepackage{listings}
\usepackage{xspace}
\usepackage{enumitem}
\usepackage{wrapfig}

\newcommand{\mc}{\mathcal}
\definecolor{mygreen}{rgb}{0.1, 0.6, 0.1}

\usepackage{titlesec}
\titlespacing\section{0pt}{4pt plus 2pt minus 2pt}{2pt plus 2pt minus 2pt}
\titlespacing\subsection{0pt}{4pt plus 4pt minus 2pt}{2pt plus 2pt minus 2pt}
\titlespacing\subsubsection{0pt}{3pt plus 4pt minus 2pt}{0pt plus 2pt minus 2pt}

\newcommand{\ours}{VLM-PC\xspace}

\definecolor{Blue9}{rgb}{0.098,0.3,0.9}
\definecolor{DarkBlue}{rgb}{0,0.08,0.45}
\usepackage{hyperref}
\hypersetup{
  colorlinks,
  urlcolor  = Blue9,
  linkcolor = Blue9,
  citecolor = Blue9,
}

\title{Commonsense Reasoning for Legged Robot Adaptation with Vision-Language Models}

\author{%
  Annie S. Chen$^*$ \\
  Stanford University \\
  \And
  Alec Lessing$^*$ \\
  Stanford University \\
  \And
  Andy Tang$^*$ \\
  Stanford University \\
  \And
  Govind Chada$^*$ \\
  Stanford University \\
  \And
  Laura Smith \\
  UC Berkeley \\
  \And
  Sergey Levine \\
  UC Berkeley \\
  \And
  Chelsea Finn \\
  Stanford University
}

\begin{document}
\maketitle
\def\thefootnote{$*$}\footnotetext{Equal Contribution. Correspondence to Annie Chen (\href{mailto:asc8@stanford.edu}{asc8@stanford.edu}). Videos of our results can be found on our website: \url{https://anniesch.github.io/vlm-pc/}.}

\begin{abstract}
    Legged robots are physically capable of navigating a diverse variety of environments and overcoming a wide range of obstructions. For example, in a search and rescue mission, a legged robot could climb over debris, crawl through gaps, and navigate out of dead ends. However, the robot's controller needs to respond intelligently to such varied obstacles, and this requires handling unexpected and unusual scenarios successfully. This presents an open challenge to current learning methods, which often struggle with generalization to the long tail of unexpected situations without heavy human supervision. To address this issue, we 
    investigate how to leverage the broad knowledge about the structure of the world and commonsense reasoning capabilities of vision-language models (VLMs) to aid legged robots in handling difficult, ambiguous
    situations. We propose a system, VLM-Predictive Control (VLM-PC), 
    combining two key components that we find to be crucial
    for eliciting on-the-fly, adaptive behavior selection with VLMs: 
    (1) in-context adaptation over previous robot interactions and (2) planning multiple skills into the future and replanning.
    We evaluate VLM-PC on several challenging real-world obstacle courses, involving dead ends and climbing and crawling, on a Go1 quadruped robot. Our experiments show that by reasoning over the history of interactions and future plans,
    VLMs enable the robot to autonomously perceive, navigate, and act in a wide range of complex
    scenarios that would otherwise require environment-specific engineering or human guidance. 
\end{abstract}


\section{Introduction}
\label{sec:intro}

Robots deployed in open-world environments must be able to handle highly unstructured and complicated environments. This is particularly the case for legged robots, which may need to operate in an extremely diverse range of circumstances. 
Consider a quadruped robot tasked with performing search and rescue in a collapsed building. This robot faces a long tail of different possible environments and obstacles, which might require climbing over debris, crawling through gaps, and backtracking and navigating out of dead ends without a map. 
Handling these diverse real-world scenarios autonomously, without detailed human guidance and specific skill directives, remains a significant challenge.
Prior work in locomotion has endowed legged robots with agile skills like running, climbing, and crawling~\citep{kuindersma2016optimization,hwangbo2019learning}, but possessing these skills alone does not solve the problem of fully autonomous deployment. To handle complex, unstructured scenarios, a robot must be able to decide how to deploy its repertoire of skills with a nuanced understanding of its situation. Consider the example of clearing a novel obstacle like debris in a collapsed building.
We expect an intelligent robot to perceive and try a skill that is likely to succeed, e.g., climbing. If the robot's attempt was unsuccessful, e.g. the debris is too slippery to climb over, the robot should recognize this and try another strategy given the information it has gathered, e.g. backtrack and try a new strategy like finding a path around the log instead.

\begin{figure*}[t]
    \centering
    \includegraphics[width=1.0\textwidth]{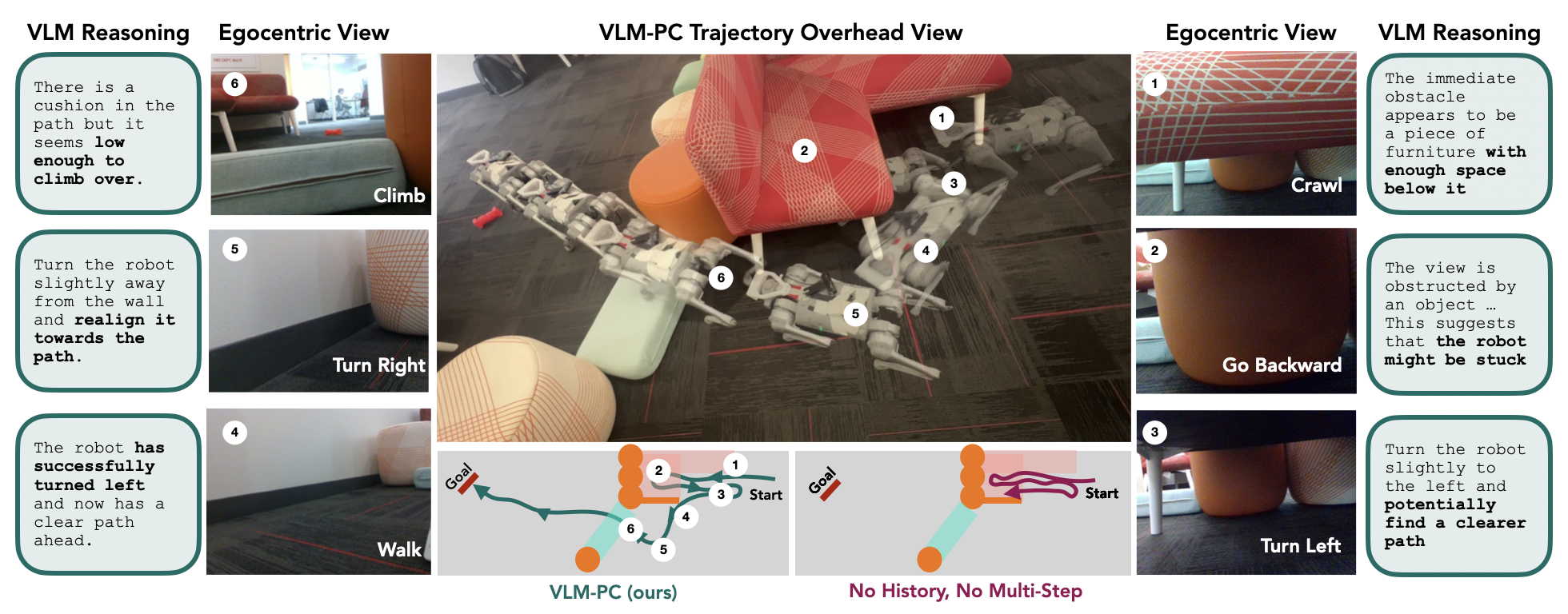}
    \vspace{-2mm}
    \caption{
        \small 
        \textbf{Vision-Language Model Predictive Control (\ours) enables real-world locomotion adaptation.} 
         By leveraging the commonsense reasoning abilities of pre-trained VLMs to adaptively select behaviors, \ours allows legged robots to quickly adjust strategies when encountering a wide range of situations, even backtracking when appropriate. \textbf{Center:} An example trajectory of the robot tasked with finding the red chew toy amid obstacles using \ours--it first crawls under a couch, then backs out of it when it finds it is a dead end, turns to walk around the couch, climbs over a sizeable cushion, and finally locates the toy. \textbf{Bottom left:} An overhead view of the trajectory with \ours. \textbf{Bottom right:} An example trajectory of the robot's behavior using a VLM naively, where the robot gets stuck and cannot adapt. \textbf{Left and right:} Visualization of the robot's egocentric POV that is provided to the VLM at different points along the trial along with excerpts of reasoning with \ours at those points.
    }
    \vspace{-7mm}
    \label{fig:teaser}
\end{figure*}

Foundation models such as vision-language models (VLMs) have the potential to help robots handle real-world, unstructured scenarios, since they possess commonsense knowledge acquired from diverse internet-scale image and language data.
Indeed, multiple prior works have shown how robots can leverage knowledge in large language models (LLMs) and VLMs for high-level planning~\cite{ahn2022can,huang2022language,driess2023palm} in robotic manipulation. In principle, VLMs can provide high-level semantic knowledge for legged robots as well, e.g. by identifying obstacles or selecting high-level behaviors. However, the use of LLMs and VLMs has been far more limited in legged robots, compared to robotic manipulation.
Furthermore, in diverse, complex environments with obstacles where the robot may get stuck and need to try multiple strategies to overcome, naively prompting a VLM to output a skill may often fail due to inaccuracies in the model's interpretation of the scene and subsequent inability to adapt with the robot's environment interactions. 

In this work, we investigate how legged robots can leverage VLMs and their general knowledge about the structure of the world and commonsense reasoning capabilities to suggest contextually informed behaviors based on visual inputs.  
We find the following two key insights crucial for facilitating adaptive behavior selection in complex, unstructured settings with VLMs:
(1) The robustness of VLMs in novel situations can be
greatly improved by taking into account the robot's interaction history, leveraging chain-of-thought reasoning~\citep{wei2022chain,kojima2022large},
and (2) Prompting the model to plan multiple skills ahead and optionally replan at each timestep is essential for foreseeing potential failures. 

Combining these insights, we propose VLM Predictive Control (\ours), which can be seen as a history-conditioned high-level analogue of visual model predictive control~\citep{garcia1989model,morari1999model,finn2017deep,ebert2018visual}
in skill space. 
With an image of the robot's view along with the history of interactions as input, the VLM is prompted to generate a multi-step plan of skills. In order to choose what plan to follow, and ultimately what skill to execute next, the model is prompted to reason through the robot's current state and whether the previous existing plan made progress on the desired task and re-plan if needed.  

In our experiments, we find that leveraging VLMs in this way allows a Go1 robot to handle a range of real-world situations that have not been tackled by prior work in a fully autonomous manner.
Across five challenging real-world settings, one of which is shown in Figure~\ref{fig:teaser}, our approach completes the target 
task around 30\% more successfully
by leveraging in-context adaptation and multi-step planning. 
Our results illustrate the potential of pre-trained VLMs, even without training on interaction data, to navigate situations autonomously from the view of a physical embodiment.

\section{Related Work}
\label{sec:related}

Our work tackles the issue of enabling legged robots to perform robustly in unstructured, unknown test-time conditions. 
Traditional model-based control approaches have achieved impressive agile locomotion~\citep{dai2014whole,kuindersma2016optimization,Hutter2016ANYmalA,Park2017BoundingCheetah,Bellicoso2018DynamicLT,Bledt2018MITC3,Katz2019MiniCA} but are not well-equipped to navigate arbitrary, open-world environments.
Learning-based approaches hold the promise of greater generalization capabilities, and training a single policy with reinforcement learning (RL) has also demonstrated successful low-level locomotion capabilities from robust walking to jumping and bipedal walking~\citep{haarnoja2018learning,tan2018sim,Yang2019DataER,yu2019sim,lee2020learning,yang2021learning,agarwal2023legged,peng2020learning,rudin2022advanced,smith2022legged,caluwaerts2023barkour,he2024agile,zhuang2023robot,cheng2023extreme}. Behind a majority of these successes is the use of domain randomization~\citep{cutler2014reinforcement,rajeswaran2016epopt,sadeghi2016cad2rl,tobin2017domain,peng2018sim,tan2018sim,yu2019sim,akkaya2019solving,xie2021dynamics,margolis2022rapid,haarnoja2023learning}, which involves training the robot under a variety of different dynamics to robustify the policy. Our work tackles an orthogonal, complementary problem: enabling legged robots to autonomously solve 
complex, partially observed tasks given a repertoire of low-level skills (which can be acquired through either traditional model-based approaches or RL training). Using these skills to solve a long-horizon task requires understanding the scene and reasoning over the information gathered in the environment, trying different low-level strategies, and adapting high-level plans on-the-fly accordingly.

Prior work has also explored utilizing a repertoire of skills to help legged robots navigate that require a combination of distinct behaviors. For example, \citet{margolis2023walk} train a policy that uses human input via remote control to select skills, while others have explored using learned models to choose appropriate behaviors on-the-fly, e.g., using search in latent space~\citep{yu2019sim,peng2020learning,yu2020learning}, direct inference using proprioceptive history~\citep{lee2020learning,kumar2021rma,fu2023deep}, prediction based on egocentric depth~\citep{miki2022learning,agarwal2023legged,zhuang2023robot,yang2023neural}, or using value functions~\citep{chen2023adapt}.
These works rely on human supervision or domain-specific information required to train model-based behavior selection. In contrast, our approach represents the robot's range of skills in language and studies how to leverage this representation with pre-trained VLMs using in-context reasoning to adapt on-the-fly in complex scenarios.

Outside legged locomotion, extensive research has explored combining prior behaviors to address long-horizon tasks, often by training high-level policies that orchestrate learned skills into complex behaviors \citep{bacon2017option,peng2019mcp,lee2019learning,sharma2020learning,strudel2020learning,nachum2018data,chitnis2020efficient,pertsch2021guided,dalal2021accelerating,nasiriany2022augmenting}.
Natural language provides a simple abstraction to index these behaviors, and using language as an abstraction for behaviors provides an interpretable space for a high-level planner to select strategies to try~\citep{ahn2022can,huang2022language,huang2022inner, macmahon2006walk,driess2023palm,misra2016tell,stepputtis2020language,kollar2010toward} or to generate robot code~\citep{liang2023code,singh2023progprompt}. 
In-context reasoning with LLMs has refined low-level behaviors~\citep{yu2023language,sha2023languagempc,mirchandani2023large,arenas2023prompt}, 
improved planning with feedback~\citep{huang2022inner} and facilitated learning from human feedback~\citep{zha2023distilling, liang2024learning}, but these do not incorporate VLMs, which can offer rich multimodal understanding.
Recent works have begun going beyond LLMs and incorporating VLMs for manipulation~\citep{huang2023voxposer, nasiriany2024pivot, belkhale2024rt} and navigation~\citep{shah2023navigation,shah2023lm}. 
Unlike these works, we focus particularly on equipping the robot to handle unpredictable situations where it might get stuck and need to explore different strategies to make progress. Enabling this in a diverse array of environments requires robust commonsense reasoning abilities, and we study the extent to which VLMs can provide these for legged robots.

While high-level planning in language grounding has been studied for manipulation or navigation tasks, it has explored far less for legged locomotion. Key works have interfaced through foot contact patterns~\citep{tang2023saytap} or code~\citep{ouyang2024long} with LLM planning. Our work implements a straightforward language-skill interface for locomotion and is the first to explore how legged robots can utilize the commonsense reasoning capabilities of pre-trained VLMs to autonomously guide adaptive behavior selection. 
In particular, the vast majority of prior works apply LLMs zero-shot based on the current instruction or observation; in this work, we study how the in-context adaptation ability of VLMs can help robots adapt to different scenarios.

\section{Problem Statement}
\label{sec:setting}

We assume the robot has access to a set of $n$ skills, which are sufficient to allow the robot to traverse the environment. Given the recent development of highly robust low-level quadrupedal locomotion controllers via RL~\cite{lee2020learning, margolis2023walk, cheng2023extreme}, we believe this assumption to be reasonable for a wide variety of real-world scenarios. For example, if the robot has the ability to climb, crawl, walk forwards and backwards, and turn in various directions, we expect that it could sequentially apply these skills to handle a variety of situations -- the challenge then is to determine when to deploy each skill to navigate an unseen, unstructured situation. 
Each skill corresponds to a policy $\pi_i$, which takes in a state $s \in \mc{S}$ and outputs a low-level action $u$. At test time, the robot interacts in a partially observed environment, where it receives images $\{I\}$ and must process them and output a skill and time duration $\delta$ that executes policy $\pi_i$ for an amount of time $\delta$. 

We frame our problem setting as an instantiation of single-life deployment~\citep{chen2022you}, where the agent has prior behaviors and is evaluated on a task during a ``single-life'' trial without any human intervention. This setting is meant to be representative of real settings in which a robot is autonomously deployed without prior knowledge of the environment or any human guidance available.
In our experimental settings with legged locomotion, this corresponds to completing a task (e.g. finding an object) by moving in a desired direction while successfully overcoming any obstacles in the terrain.

\section{Vision-Language Model Predictive Control (\ours)}
\label{sec:method}

\begin{figure*}[t!]
    \centering
    \includegraphics[width=0.99\textwidth]{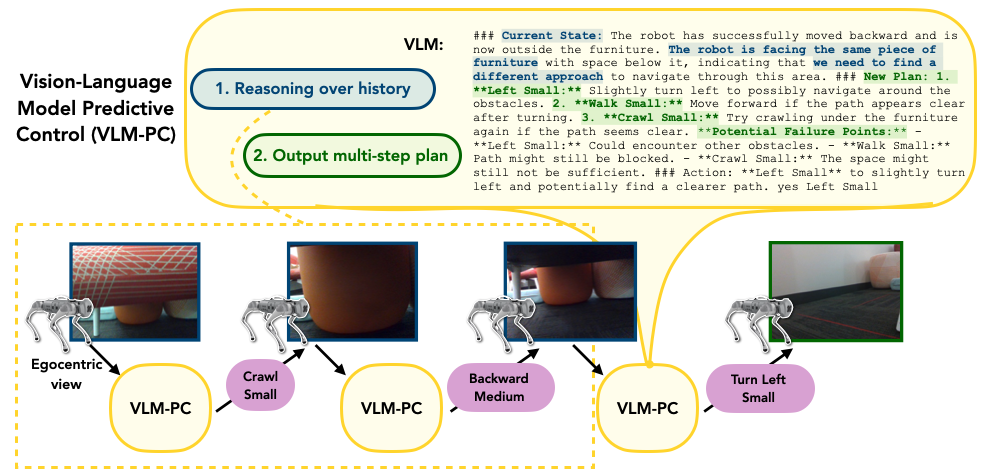}
    \caption{
        \small 
        \textbf{Vision-Language Model Predictive Control (\ours).} 
        Our method uses a pre-trained VLM to provide high-level skill commands for a legged robot to execute. 
        Given the robot's current view and history of interactions, the VLM is first prompted to reason through the robot's current state and progress with the history of commanded skills, and is then prompted to make a new multi-step plan, compare it to the prior plan, and adjust if needed. The robot executes the first skill in the plan, and the VLM is queried again. 
    }
    \label{fig:method}
\end{figure*}

Our goal is to enable legged robots to make informed decisions that lead to effective, context-aware adaptation to help navigate these situations autonomously and successfully. Our central hypothesis is that many real-world situations demand complex reasoning due to unexpected circumstances that may be difficult to generalize to. 
In this section, we first describe how we represent the robot's skills via language to be used then by a pre-trained VLM. We then detail how we prompt the VLM to reason through the robot's current state and history of interactions to select the next skill to execute to solve a task in unstructured environments.

\subsection{Interfacing Robotic Locomotion Skills with VLMs}
We consider generative VLMs, also known as multimodal language models,
which take as input $\{I, x\},$ including images $\{I\}$ and prompt text $x$ and outputs text $y$ from a distribution over textual completions $P(\cdot | \{I\}, x)$. We label each of the robot's prior behaviors $\pi_i \in \Pi$ with a command $l_i$, a textual description of the corresponding behavior. We also define levels of magnitude $m$ for each that define the duration $\delta_{l_i, m}$ that $\pi_i$ should be executed. While there are many ways to acquire locomotion policies, e.g., via traditional model-based techniques or learning-based approaches, we use the built-in controller provided by the Go1 robot. To make these policies amenable to being used by a VLM, we choose to represent the policies as skills (as opposed to less interpretable, low-level actions such as joint angles or foot contact patterns) that we acquire by varying several parameters (x- and y-velocity in the robot frame, gait type, body height, yaw speed, and duration), the details of which can be found in Appendix~\ref{sec:app-skills}. For example, $l_i$ could be ``Climb" or ``Crawl", and $m$ is ``Small'', ``Medium'', or ``Large''. At timestep $T$, the VLM is prompted to output high-level action $a_T = (l_i, m)$, which leads to the robot executing low-level actions $u_t = \pi_i(s_t; m_T)$ for the number of seconds dictated by $m_T$.
Through prompt engineering, we ensure that the VLM outputs the skill and magnitude in a specified format that allows us to extract the high-level skill command for the robot to execute.

\subsection{Using VLMs for Adaptive Behavior Selection}

We propose a system, Vision-Language Model Predictive Control (\ours), that uses a VLM to account for these errors and successively refine strategies, so that the robot can autonomously adjust from strategies that fail and try others. Summarized in Figure~\ref{fig:method}, \ours combines two key insights to effectively enable VLMs to serve as an effective high-level policy: 
(1) reasoning about information gathered by the robot in its environment and (2) selecting actions by planning ahead and iteratively replanning during execution. The VLM we use in all of our experiments is GPT-4o. We tuned the prompts to take into account the setting of legged locomotion and the limited view from the robot's camera. Full prompts and an example log of the VLM's chats are shown in Appendix~\ref{sec:app-logs}.

\paragraph{Using in-context reasoning to adapt on-the-fly.} 
In large foundation models, techniques like chain-of-thought~\citep{wei2022chain,kojima2022large,zeng2022socratic}, where the model is prompted to output intermediate reasoning steps, have been shown to significantly improve the model's ability to perform complex reasoning. 
We aim to leverage such techniques to equip the VLM to better understand and reason through the environment and provide more effective high-level commands to the robot.
In particular, we want the VLM to reason through the history in the environment and the progress made with the commanded skills before deciding on the next skill, in order to determine if the robot should try a new strategy.
As such, we include as input to the VLM an image representing the robot's current view along with the full history of interactions (including the robot's previous images and the previous outputs of the VLM) and a prompt, i.e. the input at timestep $t$ is $(I_1, x_1, y_1, I_{\delta_{m_1}}, x_{\delta_{m_1}}, y_{\delta_{m_1}}, \ldots, x_{t-1}, I_t, p_t)$, 
which contains for each previous query step $i$, each previous image $I_i$ and prompt $x_i$ along with VLM output $y_i$. 
We then prompt the VLM to first reason through what progress the robot has made using the history of commands selected and the current position and orientation of the robot. 

\paragraph{Multi-step planning and execution.}
Due to partial observability, there is often no clear answer as to which skill is most appropriate for a given situation and without physical experience, the VLM is not grounded in the robot's low-level capabilities (i.e. the VLM understands that the robot can crawl, but it does not know exactly how it crawls and whether this crawling skill will actually be successful in the current situation). So, we use an approach akin to model predictive control~\citep{garcia1989model,morari1999model,finn2017deep,ebert2018visual}, wherein we prompt the VLM to produce the immediate skill to execute by planning multiple steps $l_t$, $l_{t+\delta_m}$..., $l_{t+k}$, into the future and reasoning about the consequences of the actions. 
This allows the VLM to foresee different possible strategies that might be applicable to the current situation, so it may better adjust in the future if the next chosen skill does not make progress.
Then after the robot executes the skill corresponding to $l_t$, the VLM repeats this planning for each step during deployment.
To implement this, we specifically prompt the VLM to make a multi-step plan taking into account the latest visual observation $I_t$, compare the new plan to the prior existing plan, and use the one that seems more applicable.

\section{Experimental Results}
\label{sec:exps}

In this section, we study whether \ours can enable a Go1 quadruped robot to tackle five challenging real-world situations in a fully autonomous manner.
Concretely, we aim to answer the following empirical questions:
(1) Can \ours enable the robot to autonomously adapt in unseen, partially observed environments and effectively complete tasks that require reasoning over what strategies the robot has tried in the past?
(2) How much do in-context reasoning over the robot's experience and multi-step planning affect the robot's ability to complete these test settings?
(3) Does including additional in-context examples improve the robot's ability to handle the given setting?
We first describe our experimental setup before presenting the results of our experiments. 

\subsection{Experimental Setup}
We use a Go1 quadruped robot from Unitree.
The robot is equipped with an Intel Realsense D435 camera mounted on its head, which provides an egocentric view of the environment, which is the only source of information the robot has about its surroundings.
We configure the default controller to correspond to a set of prior behaviors: walking forward, crawling (at a low height), climbing (which can overcome stair-height obstacles), walking backward, turning left, and turning right. This same set of behaviors is used for all experiments, and details of the skills are in Appendix~\ref{sec:app-skills}. 
In each setting, we report the average and median wall clock time in seconds (where lower is better) needed to complete the task along with the success rate across five trials for each method.
If the robot does not complete the task within 100 seconds of executing actions, we consider it a failure. For each method in each setting, we report these metrics across five trials.

\begin{figure*}[t!]
    \centering
    \includegraphics[width=1.04\textwidth]{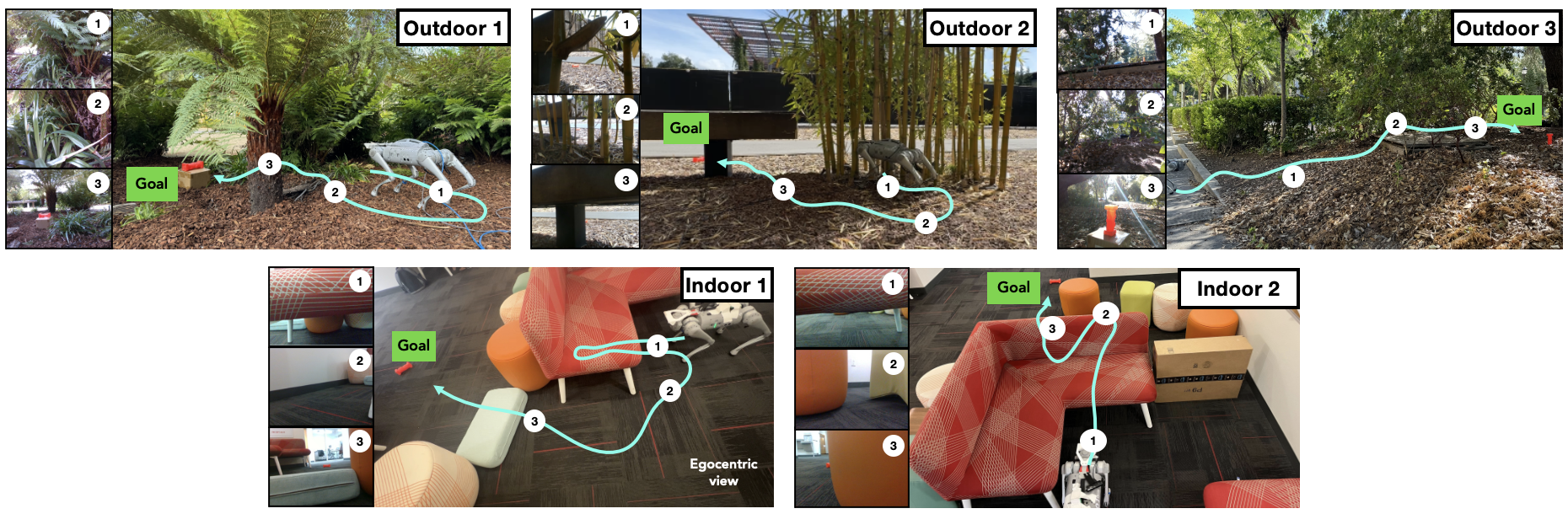}
    \caption{
        \small 
        \textbf{Deployment Environments.} We evaluate \ours on five challenging real-world settings, each of which presents unseen obstacles designed for the robot to get stuck, and requires commonsense reasoning to solve. For each setting, we give a third-person view of the obstacle course as well as an example path through the course, with three different egocentric views (labeled 1, 2, 3) at different points to show the diversity of scenes the robot encounters from its viewpoint.
    }
    \vspace{-4mm}
    \label{fig:settings}
\end{figure*}

\paragraph{Evaluation Settings.}
To evaluate each method, we conduct trials in five real-world indoor and outdoor settings. 
The settings test the robot's ability to adapt to varying terrain conditions, requiring agile skills and dynamic strategy adjustments based on new information.
The goal in each setting is to reach the ``red chew toy''.
The robot only receives information from its camera and does not have access to a map of the environment.
The tasks are shown in Figure~\ref{fig:settings}, annotated with the goal and an example path through the course, and described as follows:\\
\emph{\textbf{Indoor 1:}} The robot first must crawl under a couch, determine that it is a dead end, back up and turn to walk around the couch, climb a cushion it cannot pass without climbing, and finally locate the toy. \\
\emph{\textbf{Indoor 2:}} The robot first faces a couch that it must crawl under to the opposite side, then faces several stools blocking its path with a narrow gap between them, and must determine that it cannot fit through the gap and must turn and go around to locate the red chew toy. \\
\emph{\textbf{Outdoor 1:}} The robot first faces bushes that it must turn from and go around, then faces a series of small logs that it must climb over, and finally locates the red chew toy. \\
\emph{\textbf{Outdoor 2:}} The robot first faces a series of bamboo plants that it must turn from and go around, then a bench that it must crawl under, and then find the red chew toy.\\
\emph{\textbf{Outdoor 3:}} The robot first faces a curb that it must climb over, a dirt hill that it must walk up, a wooden plank that it must climb over, and finally locate the red chew toy between the bushes.

\paragraph{Comparisons.}
We compare \ours to several variants that differ in the amount of context provided to the VLM and the amount the VLM is prompted to plan:
(1) \textbf{No History}: The VLM is prompted with only the current image and the prompt, and is not provided with any history of interactions but is still prompted to output a multi-step plan of skills at each timestep.
(2) \textbf{No Multi-Step}: The VLM is prompted with the full history of interactions, including the robot's previous images and the previous outputs of the VLM, but is only prompted to plan a single skill at each timestep.
(3) \textbf{\ours}: The VLM is prompted with the full history of interactions, including the robot's previous images and the previous outputs of the VLM, and is prompted to make a multi-step plan of skills at each timestep.
As a baseline, we additionally compare to (4) \textbf{Random}, which randomly selects a skill and magnitude to execute at each timestep.

\subsection{Main Results}

\begin{table}[t]
\centering
\begin{subtable}[t]{\linewidth}
\resizebox{\textwidth}{!}{%
    \renewcommand{\arraystretch}{1.2}
\begin{tabular}{ccccccccccc}
\toprule
\textbf{Method} & \multicolumn{3}{c}{\textbf{Outdoor 1}} & \multicolumn{3}{c}{\textbf{Outdoor 2}} & \multicolumn{3}{c}{\textbf{Outdoor 3}} \\
\cmidrule(lr){2-4} \cmidrule(lr){5-7}  \cmidrule(lr){8-10}
 & \textbf{Avg (s) $\downarrow$} & \textbf{Median (s) $\downarrow$} & \textbf{Success (\%)} & \textbf{Avg (s) $\downarrow$} & \textbf{Median (s) $\downarrow$} & \textbf{Success (\%)} & \textbf{Avg (s) $\downarrow$} & \textbf{Median (s) $\downarrow$} & \textbf{Success (\%)} \\
\midrule
Random & 84.2 & 100 & 20 & 92 & 100 & 20 & 100 & 100 & 0 \\
No History & 82 & 100 & 20 & 100 & 100 & 0 & \textbf{49.6} & \textbf{17.1} & \textbf{60} \\
No Multi-Step & 81.5 & 100 & 20 & 100 & 100 & 0 & 57.4 & 42 & \textbf{60} \\
VLM-PC & \textbf{49.4} & \textbf{17} & \textbf{60} & \textbf{68.8} & \textbf{65.5} & \textbf{60} & 61.7 & 50.5 & \textbf{60} \\
\bottomrule
\end{tabular}
} 
\end{subtable}
\begin{subtable}[t]{0.67\linewidth}
\resizebox{1\textwidth}{!}{%
    \renewcommand{\arraystretch}{1.2}
\begin{tabular}{ccccccccccc}
\toprule
\textbf{Method} & \multicolumn{3}{c}{\textbf{Indoor 1}} & \multicolumn{3}{c}{\textbf{Indoor 2}} \\
\cmidrule(lr){2-4} \cmidrule(lr){5-7}
 & \textbf{Avg (s) $\downarrow$} & \textbf{Median (s) $\downarrow$} & \textbf{Success (\%)} & \textbf{Avg (s) $\downarrow$} & \textbf{Median (s) $\downarrow$} & \textbf{Success (\%)} \\
\midrule
Random & 100 & 100 & 0 & 93.9 & 100 & 20 \\
No History & 100 & 100 & 0 & 87.9 & 100 & 20 \\
No Multi-Step & \textbf{57.2} & \textbf{34} & \textbf{60} & 82.9 & 100 & 40 \\
VLM-PC & 66.7 & 46.7 & \textbf{60} & \textbf{37.1} & \textbf{35.3} & \textbf{80} \\
\bottomrule
\end{tabular}
}
\end{subtable}
\vspace{3mm}
\caption{\small \textbf{Results on Each Setting.} We report the average and median time to complete the task (where lower is better) and success rate, across five trials for each method in each of the five settings. \ours far outperforms the comparisons across all metrics in three of the scenes (Outdoor 1, Outdoor 2, and Indoor 2) and is comparable to the best other method in the other two scenes. Furthermore, \ours is the only method that succeeds a majority of the time in each setting. }
\label{tab:results}
\end{table}

\begin{figure*}[t!]
    \centering
    \includegraphics[width=1.02\textwidth]{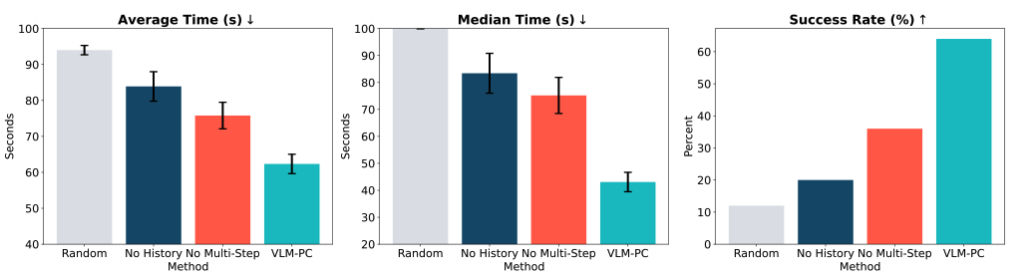}
    \caption{
        \small 
        \textbf{Main Results Averaged Across Settings.}  
        Across all five settings on average, \ours significantly outperforms Random, No History, and No Multi-Step on average and median time to complete the task and success rate, performing roughly 30\% more successfully than the next best method. 
    }
    \vspace{-4mm}
    \label{fig:avg-results}
\end{figure*}

\begin{figure*}[t!]
    \centering
    \includegraphics[width=1.01\textwidth]{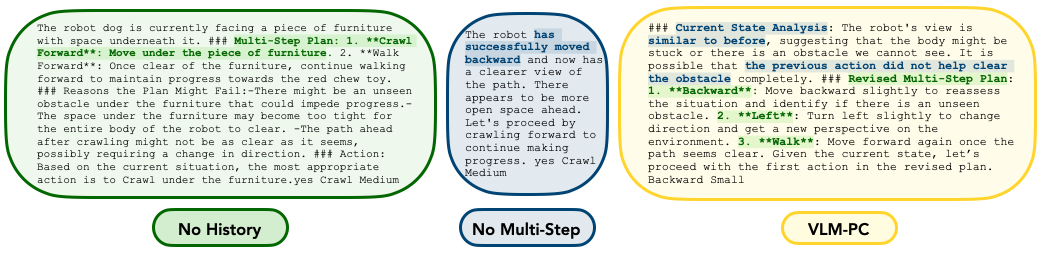}
    \caption{
        \small 
        \textbf{Typical VLM Interactions.} With \ours (Right), the VLM can both analyze the efficacy of previous commands and prepare new, coherent plans to tackle the current obstacle, by combining benefits from multi-step planning from No History (Left) and reasoning over history from No Multi-Step (Center).
    }
    \label{fig:qual-analysis}
\end{figure*}

As shown in Figure~\ref{fig:avg-results}, on average across all five settings, \ours successfully completes the task 64\% of the time, almost 30\% more than the second best method (No Multi-Step), which succeeds on average 36\% of the time. \ours is also on average over 20\% faster at completing the target task as the next best method, showing that including both history and multi-step planning are important for improving the use of these VLMs in providing high-level commands in a variety of settings. In Figure~\ref{fig:settings}, we find that particularly on Indoor 2, Outdoor 1, and Outdoor 2, \ours is more than twice as successful as the next best method. No Multi-Step is the second best method, and does comparably to \ours (which does multi-step planning) on Indoor 1 and Outdoor 3, indicating that in some situations, multi-step planning does not significantly help, although it does not hurt performance. No History fails in almost every setting except Outdoor 3, as it often gets stuck behind obstacles that require trying multiple different strategies. Random fails in every setting, showing that each setting requires nontrivial reasoning for the robot to succeed. 
We provide examples of typical interactions with the VLM in Figure~\ref{fig:qual-analysis} without history, without multi-step planning, compared to \ours, and we find that with ours, the VLM is able to both reason through effectiveness of prior strategies and plan ahead to try coherent new strategies to overcome the current obstacle.

\subsection{Adding Labeled In-Context Examples Can Improve Performance}
\begin{wrapfigure}{r}{0.6\textwidth}
    \centering
    \begin{adjustbox}{width=0.58\textwidth}
        \begin{tabular}{llccc}
            \toprule
            Course    & Method       & Avg time (s) & Median time (s) & Success Rate (\%) \\
            \midrule
            Indoor 2  & VLM-PC       & 37.1          & 35.3             & 80                \\
                      & VLM-PC + IC  & 13.5        & 13.7            & 100               \\
            \midrule
            Outdoor 1 & VLM-PC       & 49.4         & 17.0              & 60                \\
                      & VLM-PC + IC  & 10.0           & 10.0              & 100               \\
            \bottomrule
        \end{tabular}
    \end{adjustbox}
    \caption{\small \textbf{VLM-PC with Labeled In-Context Examples.} We find that in two of the obstacle courses, leveraging the VLM's in-context learning capabilities by providing additional images labeled with the best command can significantly improve performance.}
    \label{tab:icl}
\end{wrapfigure}

As large foundation models trained on Internet-scale data are used as these high level planners, they can leverage in-context learning, where examples or instructions are included as context in the input to the model~\citep{min2022rethinking,dong2022survey}.
We provide an extension of our method including in-context examples, called VLM-PC+IC, where we include in the first prompt several additional images, taken from the egocentric view at different points in the environment, as well as a label for each of them with the best command to take.
This provides the VLM with more context about the environment and the best strategies to take at key points.
As shown in Table~\ref{tab:icl}, we find that in two of the obstacle courses, this can significantly improve performance. 
While inexpensive to obtain, this does require human labeling of several images from the deployment environment with the best command to take, which may not be feasible in all deployment settings.
Nonetheless, this extension further reinforces the importance of providing useful context to the VLM and having it use this context to make informed decisions, and shows that these labeled examples can be useful context on top of the history of experiences in the environment.

\section{Discussion and Limitations}
\label{sec:conclusion}
We introduced Vision-Language Model Predictive Control (\ours), which enables legged robots to rapidly adapt to changing, unseen circumstances during deployment. 
On a Go1 quadruped robot, we find that \ours can autonomously handle a range of complex real-world tasks involving climbing over obstacles, crawling under furniture, and navigating around dead ends and through cluttered environments.
While \ours is promising solution for enabling legged robots to handle new tasks, there remains much left to explore regarding how to best leverage VLMs for adaptive behavior for legged robots, especially as core VLM capabilities continue to improve.
First, improving language grounding for locomotion to better capture the nuances of the robot's capabilities could lead to more effective decision-making. It would also be interesting to explore if fine-tuning these VLMs, perhaps with techniques like reinforcement learning from human feedback, can lead to more efficient reasoning.
In addition, the camera is currently mounted on the head of the robot and provides a limited field of view, making it challenging for the agent to understand the full context of its environment.
This limitation sometimes causes the VLM to struggle with scenarios where the robot's body may be stuck on an obstacle even if the head has cleared it.
Incorporating more sensors or scene reconstruction could provide a more comprehensive view of the environment, allowing the VLM to reason more effectively.
Finally, it would be interesting to explore how to use VLMs to combine high-level planning for locomotion with that for manipulation, to enable robots to handle a wider range of tasks.

\acknowledgments{We thank Zipeng Fu and Yunhao Cao for help with hardware, and Jonathan Yang, Anikait Singh, and other members of the IRIS lab for helpful discussions on this project.
This work was supported by an NSF CAREER award, ONR grant N00014-21-1-2685 and N00014-22-1-2621, Volkswagen, and the AI Institute, and an NSF graduate research fellowship.}

\bibliography{example}  %

\begin{thebibliography}{81}
\providecommand{\natexlab}[1]{#1}
\providecommand{\url}[1]{\texttt{#1}}
\expandafter\ifx\csname urlstyle\endcsname\relax
  \providecommand{\doi}[1]{doi: #1}\else
  \providecommand{\doi}{doi: \begingroup \urlstyle{rm}\Url}\fi

\bibitem[Kuindersma et~al.(2016)Kuindersma, Deits, Fallon, Valenzuela, Dai, Permenter, Koolen, Marion, and Tedrake]{kuindersma2016optimization}
S.~Kuindersma, R.~Deits, M.~Fallon, A.~Valenzuela, H.~Dai, F.~Permenter, T.~Koolen, P.~Marion, and R.~Tedrake.
\newblock Optimization-based locomotion planning, estimation, and control design for the atlas humanoid robot.
\newblock \emph{Autonomous robots}, 40:\penalty0 429--455, 2016.

\bibitem[Hwangbo et~al.(2019)Hwangbo, Lee, Dosovitskiy, Bellicoso, Tsounis, Koltun, and Hutter]{hwangbo2019learning}
J.~Hwangbo, J.~Lee, A.~Dosovitskiy, D.~Bellicoso, V.~Tsounis, V.~Koltun, and M.~Hutter.
\newblock Learning agile and dynamic motor skills for legged robots.
\newblock \emph{Science Robotics}, 4\penalty0 (26):\penalty0 eaau5872, 2019.

\bibitem[Ahn et~al.(2022)Ahn, Brohan, Brown, Chebotar, Cortes, David, Finn, Fu, Gopalakrishnan, Hausman, et~al.]{ahn2022can}
M.~Ahn, A.~Brohan, N.~Brown, Y.~Chebotar, O.~Cortes, B.~David, C.~Finn, C.~Fu, K.~Gopalakrishnan, K.~Hausman, et~al.
\newblock Do as i can, not as i say: Grounding language in robotic affordances.
\newblock \emph{arXiv preprint arXiv:2204.01691}, 2022.

\bibitem[Huang et~al.(2022)Huang, Abbeel, Pathak, and Mordatch]{huang2022language}
W.~Huang, P.~Abbeel, D.~Pathak, and I.~Mordatch.
\newblock Language models as zero-shot planners: Extracting actionable knowledge for embodied agents.
\newblock In \emph{International Conference on Machine Learning}, pages 9118--9147. PMLR, 2022.

\bibitem[Driess et~al.(2023)Driess, Xia, Sajjadi, Lynch, Chowdhery, Ichter, Wahid, Tompson, Vuong, Yu, et~al.]{driess2023palm}
D.~Driess, F.~Xia, M.~S. Sajjadi, C.~Lynch, A.~Chowdhery, B.~Ichter, A.~Wahid, J.~Tompson, Q.~Vuong, T.~Yu, et~al.
\newblock Palm-e: An embodied multimodal language model.
\newblock \emph{arXiv preprint arXiv:2303.03378}, 2023.

\bibitem[Wei et~al.(2022)Wei, Wang, Schuurmans, Bosma, Xia, Chi, Le, Zhou, et~al.]{wei2022chain}
J.~Wei, X.~Wang, D.~Schuurmans, M.~Bosma, F.~Xia, E.~Chi, Q.~V. Le, D.~Zhou, et~al.
\newblock Chain-of-thought prompting elicits reasoning in large language models.
\newblock \emph{Advances in neural information processing systems}, 35:\penalty0 24824--24837, 2022.

\bibitem[Kojima et~al.(2022)Kojima, Gu, Reid, Matsuo, and Iwasawa]{kojima2022large}
T.~Kojima, S.~S. Gu, M.~Reid, Y.~Matsuo, and Y.~Iwasawa.
\newblock Large language models are zero-shot reasoners.
\newblock \emph{Advances in neural information processing systems}, 35:\penalty0 22199--22213, 2022.

\bibitem[Garcia et~al.(1989)Garcia, Prett, and Morari]{garcia1989model}
C.~E. Garcia, D.~M. Prett, and M.~Morari.
\newblock Model predictive control: Theory and practice—a survey.
\newblock \emph{Automatica}, 25\penalty0 (3):\penalty0 335--348, 1989.

\bibitem[Morari and Lee(1999)]{morari1999model}
M.~Morari and J.~H. Lee.
\newblock Model predictive control: past, present and future.
\newblock \emph{Computers \& chemical engineering}, 23\penalty0 (4-5):\penalty0 667--682, 1999.

\bibitem[Finn and Levine(2017)]{finn2017deep}
C.~Finn and S.~Levine.
\newblock Deep visual foresight for planning robot motion.
\newblock In \emph{2017 IEEE International Conference on Robotics and Automation (ICRA)}, pages 2786--2793. IEEE, 2017.

\bibitem[Ebert et~al.(2018)Ebert, Finn, Dasari, Xie, Lee, and Levine]{ebert2018visual}
F.~Ebert, C.~Finn, S.~Dasari, A.~Xie, A.~Lee, and S.~Levine.
\newblock Visual foresight: Model-based deep reinforcement learning for vision-based robotic control.
\newblock \emph{arXiv preprint arXiv:1812.00568}, 2018.

\bibitem[Dai et~al.(2014)Dai, Valenzuela, and Tedrake]{dai2014whole}
H.~Dai, A.~Valenzuela, and R.~Tedrake.
\newblock Whole-body motion planning with centroidal dynamics and full kinematics.
\newblock In \emph{2014 IEEE-RAS International Conference on Humanoid Robots}, pages 295--302. IEEE, 2014.

\bibitem[Hutter et~al.(2016)Hutter, Gehring, Jud, Lauber, Bellicoso, Tsounis, Hwangbo, Bodie, Fankhauser, Bloesch, Diethelm, Bachmann, Melzer, and H{\"o}pflinger]{Hutter2016ANYmalA}
M.~Hutter, C.~Gehring, D.~Jud, A.~Lauber, D.~Bellicoso, V.~Tsounis, J.~Hwangbo, K.~Bodie, P.~Fankhauser, M.~Bloesch, R.~Diethelm, S.~Bachmann, A.~Melzer, and M.~H{\"o}pflinger.
\newblock Anymal - a highly mobile and dynamic quadrupedal robot.
\newblock pages 38--44, 2016.

\bibitem[Park et~al.(2017)Park, Wensing, and Kim]{Park2017BoundingCheetah}
H.-W. Park, P.~M. Wensing, and S.~Kim.
\newblock High-speed bounding with the mit cheetah 2: Control design and experiments.
\newblock \emph{The International Journal of Robotics Research}, 36\penalty0 (2):\penalty0 167--192, 2017.
\newblock \doi{10.1177/0278364917694244}.
\newblock URL \url{https://doi.org/10.1177/0278364917694244}.

\bibitem[Bellicoso et~al.(2018)Bellicoso, Jenelten, Gehring, and Hutter]{Bellicoso2018DynamicLT}
D.~Bellicoso, F.~Jenelten, C.~Gehring, and M.~Hutter.
\newblock Dynamic locomotion through online nonlinear motion optimization for quadrupedal robots.
\newblock \emph{IEEE Robotics and Automation Letters}, 3:\penalty0 2261--2268, 2018.

\bibitem[Bledt et~al.(2018)Bledt, Powell, Katz, Carlo, Wensing, and Kim]{Bledt2018MITC3}
G.~Bledt, M.~J. Powell, B.~Katz, J.~Carlo, P.~Wensing, and S.~Kim.
\newblock Mit cheetah 3: Design and control of a robust, dynamic quadruped robot.
\newblock pages 2245--2252, 2018.

\bibitem[Katz et~al.(2019)Katz, Carlo, and Kim]{Katz2019MiniCA}
B.~Katz, J.~Carlo, and S.~Kim.
\newblock Mini cheetah: A platform for pushing the limits of dynamic quadruped control.
\newblock \emph{2019 International Conference on Robotics and Automation (ICRA)}, pages 6295--6301, 2019.

\bibitem[Haarnoja et~al.(2018)Haarnoja, Ha, Zhou, Tan, Tucker, and Levine]{haarnoja2018learning}
T.~Haarnoja, S.~Ha, A.~Zhou, J.~Tan, G.~Tucker, and S.~Levine.
\newblock Learning to walk via deep reinforcement learning.
\newblock \emph{arXiv preprint arXiv:1812.11103}, 2018.

\bibitem[Tan et~al.(2018)Tan, Zhang, Coumans, Iscen, Bai, Hafner, Bohez, and Vanhoucke]{tan2018sim}
J.~Tan, T.~Zhang, E.~Coumans, A.~Iscen, Y.~Bai, D.~Hafner, S.~Bohez, and V.~Vanhoucke.
\newblock Sim-to-real: Learning agile locomotion for quadruped robots.
\newblock \emph{arXiv preprint arXiv:1804.10332}, 2018.

\bibitem[Yang et~al.(2019)Yang, Caluwaerts, Iscen, Zhang, Tan, and Sindhwani]{Yang2019DataER}
Y.~Yang, K.~Caluwaerts, A.~Iscen, T.~Zhang, J.~Tan, and V.~Sindhwani.
\newblock Data efficient reinforcement learning for legged robots.
\newblock \emph{Conference on Robot Learning}, abs/1907.03613, 2019.

\bibitem[Yu et~al.(2019)Yu, Kumar, Turk, and Liu]{yu2019sim}
W.~Yu, V.~C. Kumar, G.~Turk, and C.~K. Liu.
\newblock Sim-to-real transfer for biped locomotion.
\newblock In \emph{2019 ieee/rsj international conference on intelligent robots and systems (iros)}, pages 3503--3510. IEEE, 2019.

\bibitem[Lee et~al.(2020)Lee, Hwangbo, Wellhausen, Koltun, and Hutter]{lee2020learning}
J.~Lee, J.~Hwangbo, L.~Wellhausen, V.~Koltun, and M.~Hutter.
\newblock Learning quadrupedal locomotion over challenging terrain.
\newblock \emph{Science robotics}, 5\penalty0 (47):\penalty0 eabc5986, 2020.

\bibitem[Yang et~al.(2021)Yang, Zhang, Hansen, Xu, and Wang]{yang2021learning}
R.~Yang, M.~Zhang, N.~Hansen, H.~Xu, and X.~Wang.
\newblock Learning vision-guided quadrupedal locomotion end-to-end with cross-modal transformers.
\newblock \emph{arXiv preprint arXiv:2107.03996}, 2021.

\bibitem[Agarwal et~al.(2022)Agarwal, Kumar, Malik, and Pathak]{agarwal2023legged}
A.~Agarwal, A.~Kumar, J.~Malik, and D.~Pathak.
\newblock Legged locomotion in challenging terrains using egocentric vision.
\newblock In \emph{Conference on Robot Learning}, 2022.

\bibitem[Peng et~al.(2020)Peng, Coumans, Zhang, Lee, Tan, and Levine]{peng2020learning}
X.~B. Peng, E.~Coumans, T.~Zhang, T.-W. Lee, J.~Tan, and S.~Levine.
\newblock Learning agile robotic locomotion skills by imitating animals.
\newblock \emph{arXiv preprint arXiv:2004.00784}, 2020.

\bibitem[Rudin et~al.(2022)Rudin, Hoeller, Bjelonic, and Hutter]{rudin2022advanced}
N.~Rudin, D.~Hoeller, M.~Bjelonic, and M.~Hutter.
\newblock Advanced skills by learning locomotion and local navigation end-to-end.
\newblock In \emph{2022 IEEE/RSJ International Conference on Intelligent Robots and Systems (IROS)}, pages 2497--2503. IEEE, 2022.

\bibitem[Smith et~al.(2022)Smith, Kew, Peng, Ha, Tan, and Levine]{smith2022legged}
L.~Smith, J.~C. Kew, X.~B. Peng, S.~Ha, J.~Tan, and S.~Levine.
\newblock Legged robots that keep on learning: Fine-tuning locomotion policies in the real world.
\newblock In \emph{2022 International Conference on Robotics and Automation (ICRA)}, pages 1593--1599. IEEE, 2022.

\bibitem[Caluwaerts et~al.(2023)Caluwaerts, Iscen, Kew, Yu, Zhang, Freeman, Lee, Lee, Saliceti, Zhuang, et~al.]{caluwaerts2023barkour}
K.~Caluwaerts, A.~Iscen, J.~C. Kew, W.~Yu, T.~Zhang, D.~Freeman, K.-H. Lee, L.~Lee, S.~Saliceti, V.~Zhuang, et~al.
\newblock Barkour: Benchmarking animal-level agility with quadruped robots.
\newblock \emph{arXiv preprint arXiv:2305.14654}, 2023.

\bibitem[He et~al.(2024)He, Zhang, Xiao, He, Liu, and Shi]{he2024agile}
T.~He, C.~Zhang, W.~Xiao, G.~He, C.~Liu, and G.~Shi.
\newblock Agile but safe: Learning collision-free high-speed legged locomotion.
\newblock \emph{arXiv preprint arXiv:2401.17583}, 2024.

\bibitem[Zhuang et~al.(2023)Zhuang, Fu, Wang, Atkeson, Schwertfeger, Finn, and Zhao]{zhuang2023robot}
Z.~Zhuang, Z.~Fu, J.~Wang, C.~G. Atkeson, S.~Schwertfeger, C.~Finn, and H.~Zhao.
\newblock Robot parkour learning.
\newblock In \emph{7th Annual Conference on Robot Learning}, 2023.

\bibitem[Cheng et~al.(2023)Cheng, Shi, Agarwal, and Pathak]{cheng2023extreme}
X.~Cheng, K.~Shi, A.~Agarwal, and D.~Pathak.
\newblock Extreme parkour with legged robots.
\newblock \emph{arXiv preprint arXiv:2309.14341}, 2023.

\bibitem[Cutler et~al.(2014)Cutler, Walsh, and How]{cutler2014reinforcement}
M.~Cutler, T.~J. Walsh, and J.~P. How.
\newblock Reinforcement learning with multi-fidelity simulators.
\newblock In \emph{2014 IEEE International Conference on Robotics and Automation (ICRA)}, pages 3888--3895. IEEE, 2014.

\bibitem[Rajeswaran et~al.(2016)Rajeswaran, Ghotra, Ravindran, and Levine]{rajeswaran2016epopt}
A.~Rajeswaran, S.~Ghotra, B.~Ravindran, and S.~Levine.
\newblock Epopt: Learning robust neural network policies using model ensembles.
\newblock \emph{arXiv preprint arXiv:1610.01283}, 2016.

\bibitem[Sadeghi and Levine(2016)]{sadeghi2016cad2rl}
F.~Sadeghi and S.~Levine.
\newblock Cad2rl: Real single-image flight without a single real image.
\newblock \emph{arXiv preprint arXiv:1611.04201}, 2016.

\bibitem[Tobin et~al.(2017)Tobin, Fong, Ray, Schneider, Zaremba, and Abbeel]{tobin2017domain}
J.~Tobin, R.~Fong, A.~Ray, J.~Schneider, W.~Zaremba, and P.~Abbeel.
\newblock Domain randomization for transferring deep neural networks from simulation to the real world.
\newblock In \emph{2017 IEEE/RSJ international conference on intelligent robots and systems (IROS)}, pages 23--30. IEEE, 2017.

\bibitem[Peng et~al.(2018)Peng, Andrychowicz, Zaremba, and Abbeel]{peng2018sim}
X.~B. Peng, M.~Andrychowicz, W.~Zaremba, and P.~Abbeel.
\newblock Sim-to-real transfer of robotic control with dynamics randomization.
\newblock In \emph{2018 IEEE international conference on robotics and automation (ICRA)}, pages 3803--3810. IEEE, 2018.

\bibitem[Akkaya et~al.(2019)Akkaya, Andrychowicz, Chociej, Litwin, McGrew, Petron, Paino, Plappert, Powell, Ribas, et~al.]{akkaya2019solving}
I.~Akkaya, M.~Andrychowicz, M.~Chociej, M.~Litwin, B.~McGrew, A.~Petron, A.~Paino, M.~Plappert, G.~Powell, R.~Ribas, et~al.
\newblock Solving rubik's cube with a robot hand.
\newblock \emph{arXiv preprint arXiv:1910.07113}, 2019.

\bibitem[Xie et~al.(2021)Xie, Da, Van~de Panne, Babich, and Garg]{xie2021dynamics}
Z.~Xie, X.~Da, M.~Van~de Panne, B.~Babich, and A.~Garg.
\newblock Dynamics randomization revisited: A case study for quadrupedal locomotion.
\newblock In \emph{2021 IEEE International Conference on Robotics and Automation (ICRA)}, pages 4955--4961. IEEE, 2021.

\bibitem[Margolis et~al.(2022)Margolis, Yang, Paigwar, Chen, and Agrawal]{margolis2022rapid}
G.~B. Margolis, G.~Yang, K.~Paigwar, T.~Chen, and P.~Agrawal.
\newblock Rapid locomotion via reinforcement learning.
\newblock \emph{arXiv preprint arXiv:2205.02824}, 2022.

\bibitem[Haarnoja et~al.(2023)Haarnoja, Moran, Lever, Huang, Tirumala, Wulfmeier, Humplik, Tunyasuvunakool, Siegel, Hafner, et~al.]{haarnoja2023learning}
T.~Haarnoja, B.~Moran, G.~Lever, S.~H. Huang, D.~Tirumala, M.~Wulfmeier, J.~Humplik, S.~Tunyasuvunakool, N.~Y. Siegel, R.~Hafner, et~al.
\newblock Learning agile soccer skills for a bipedal robot with deep reinforcement learning.
\newblock \emph{arXiv preprint arXiv:2304.13653}, 2023.

\bibitem[Margolis and Agrawal(2023)]{margolis2023walk}
G.~B. Margolis and P.~Agrawal.
\newblock Walk these ways: Tuning robot control for generalization with multiplicity of behavior.
\newblock In \emph{Conference on Robot Learning}, pages 22--31. PMLR, 2023.

\bibitem[Yu et~al.(2020)Yu, Tan, Bai, Coumans, and Ha]{yu2020learning}
W.~Yu, J.~Tan, Y.~Bai, E.~Coumans, and S.~Ha.
\newblock Learning fast adaptation with meta strategy optimization.
\newblock \emph{IEEE Robotics and Automation Letters}, 2020.

\bibitem[Kumar et~al.(2021)Kumar, Fu, Pathak, and Malik]{kumar2021rma}
A.~Kumar, Z.~Fu, D.~Pathak, and J.~Malik.
\newblock Rma: Rapid motor adaptation for legged robots.
\newblock \emph{arXiv preprint arXiv:2107.04034}, 2021.

\bibitem[Fu et~al.(2022)Fu, Cheng, and Pathak]{fu2023deep}
Z.~Fu, X.~Cheng, and D.~Pathak.
\newblock Deep whole-body control: learning a unified policy for manipulation and locomotion.
\newblock In \emph{Conference on Robot Learning}, 2022.

\bibitem[Miki et~al.(2022)Miki, Lee, Hwangbo, Wellhausen, Koltun, and Hutter]{miki2022learning}
T.~Miki, J.~Lee, J.~Hwangbo, L.~Wellhausen, V.~Koltun, and M.~Hutter.
\newblock Learning robust perceptive locomotion for quadrupedal robots in the wild.
\newblock \emph{Science Robotics}, 2022.

\bibitem[Yang et~al.(2023)Yang, Yang, and Wang]{yang2023neural}
R.~Yang, G.~Yang, and X.~Wang.
\newblock Neural volumetric memory for visual locomotion control.
\newblock In \emph{Proceedings of the IEEE/CVF Conference on Computer Vision and Pattern Recognition}, 2023.

\bibitem[Chen et~al.(2023)Chen, Chada, Smith, Sharma, Fu, Levine, and Finn]{chen2023adapt}
A.~S. Chen, G.~Chada, L.~Smith, A.~Sharma, Z.~Fu, S.~Levine, and C.~Finn.
\newblock Adapt on-the-go: Behavior modulation for single-life robot deployment.
\newblock \emph{arXiv preprint arXiv:2311.01059}, 2023.

\bibitem[Bacon et~al.(2017)Bacon, Harb, and Precup]{bacon2017option}
P.-L. Bacon, J.~Harb, and D.~Precup.
\newblock The option-critic architecture.
\newblock In \emph{Proceedings of the AAAI conference on artificial intelligence}, volume~31, 2017.

\bibitem[Peng et~al.(2019)Peng, Chang, Zhang, Abbeel, and Levine]{peng2019mcp}
X.~B. Peng, M.~Chang, G.~Zhang, P.~Abbeel, and S.~Levine.
\newblock Mcp: Learning composable hierarchical control with multiplicative compositional policies.
\newblock \emph{Advances in Neural Information Processing Systems}, 32, 2019.

\bibitem[Lee et~al.(2019)Lee, Yang, and Lim]{lee2019learning}
Y.~Lee, J.~Yang, and J.~J. Lim.
\newblock Learning to coordinate manipulation skills via skill behavior diversification.
\newblock In \emph{International conference on learning representations}, 2019.

\bibitem[Sharma et~al.(2020)Sharma, Liang, Zhao, LaGrassa, and Kroemer]{sharma2020learning}
M.~Sharma, J.~Liang, J.~Zhao, A.~LaGrassa, and O.~Kroemer.
\newblock Learning to compose hierarchical object-centric controllers for robotic manipulation.
\newblock \emph{arXiv preprint arXiv:2011.04627}, 2020.

\bibitem[Strudel et~al.(2020)Strudel, Pashevich, Kalevatykh, Laptev, Sivic, and Schmid]{strudel2020learning}
R.~Strudel, A.~Pashevich, I.~Kalevatykh, I.~Laptev, J.~Sivic, and C.~Schmid.
\newblock Learning to combine primitive skills: A step towards versatile robotic manipulation.
\newblock In \emph{2020 IEEE International Conference on Robotics and Automation (ICRA)}, pages 4637--4643. IEEE, 2020.

\bibitem[Nachum et~al.(2018)Nachum, Gu, Lee, and Levine]{nachum2018data}
O.~Nachum, S.~S. Gu, H.~Lee, and S.~Levine.
\newblock Data-efficient hierarchical reinforcement learning.
\newblock \emph{Advances in neural information processing systems}, 31, 2018.

\bibitem[Chitnis et~al.(2020)Chitnis, Tulsiani, Gupta, and Gupta]{chitnis2020efficient}
R.~Chitnis, S.~Tulsiani, S.~Gupta, and A.~Gupta.
\newblock Efficient bimanual manipulation using learned task schemas.
\newblock In \emph{2020 IEEE International Conference on Robotics and Automation (ICRA)}, pages 1149--1155. IEEE, 2020.

\bibitem[Pertsch et~al.(2021)Pertsch, Lee, Wu, and Lim]{pertsch2021guided}
K.~Pertsch, Y.~Lee, Y.~Wu, and J.~J. Lim.
\newblock Guided reinforcement learning with learned skills.
\newblock \emph{arXiv preprint arXiv:2107.10253}, 2021.

\bibitem[Dalal et~al.(2021)Dalal, Pathak, and Salakhutdinov]{dalal2021accelerating}
M.~Dalal, D.~Pathak, and R.~R. Salakhutdinov.
\newblock Accelerating robotic reinforcement learning via parameterized action primitives.
\newblock \emph{Advances in Neural Information Processing Systems}, 34:\penalty0 21847--21859, 2021.

\bibitem[Nasiriany et~al.(2022)Nasiriany, Liu, and Zhu]{nasiriany2022augmenting}
S.~Nasiriany, H.~Liu, and Y.~Zhu.
\newblock Augmenting reinforcement learning with behavior primitives for diverse manipulation tasks.
\newblock In \emph{2022 International Conference on Robotics and Automation (ICRA)}, pages 7477--7484. IEEE, 2022.

\bibitem[Huang et~al.(2022)Huang, Xia, Xiao, Chan, Liang, Florence, Zeng, Tompson, Mordatch, Chebotar, et~al.]{huang2022inner}
W.~Huang, F.~Xia, T.~Xiao, H.~Chan, J.~Liang, P.~Florence, A.~Zeng, J.~Tompson, I.~Mordatch, Y.~Chebotar, et~al.
\newblock Inner monologue: Embodied reasoning through planning with language models.
\newblock \emph{arXiv preprint arXiv:2207.05608}, 2022.

\bibitem[MacMahon et~al.(2006)MacMahon, Stankiewicz, and Kuipers]{macmahon2006walk}
M.~MacMahon, B.~Stankiewicz, and B.~Kuipers.
\newblock Walk the talk: Connecting language, knowledge, and action in route instructions.
\newblock \emph{Def}, 2\penalty0 (6):\penalty0 4, 2006.

\bibitem[Misra et~al.(2016)Misra, Sung, Lee, and Saxena]{misra2016tell}
D.~K. Misra, J.~Sung, K.~Lee, and A.~Saxena.
\newblock Tell me dave: Context-sensitive grounding of natural language to manipulation instructions.
\newblock \emph{The International Journal of Robotics Research}, 35\penalty0 (1-3):\penalty0 281--300, 2016.

\bibitem[Stepputtis et~al.(2020)Stepputtis, Campbell, Phielipp, Lee, Baral, and Ben~Amor]{stepputtis2020language}
S.~Stepputtis, J.~Campbell, M.~Phielipp, S.~Lee, C.~Baral, and H.~Ben~Amor.
\newblock Language-conditioned imitation learning for robot manipulation tasks.
\newblock \emph{Advances in Neural Information Processing Systems}, 33:\penalty0 13139--13150, 2020.

\bibitem[Kollar et~al.(2010)Kollar, Tellex, Roy, and Roy]{kollar2010toward}
T.~Kollar, S.~Tellex, D.~Roy, and N.~Roy.
\newblock Toward understanding natural language directions.
\newblock In \emph{2010 5th ACM/IEEE International Conference on Human-Robot Interaction (HRI)}, pages 259--266. IEEE, 2010.

\bibitem[Liang et~al.(2023)Liang, Huang, Xia, Xu, Hausman, Ichter, Florence, and Zeng]{liang2023code}
J.~Liang, W.~Huang, F.~Xia, P.~Xu, K.~Hausman, B.~Ichter, P.~Florence, and A.~Zeng.
\newblock Code as policies: Language model programs for embodied control.
\newblock In \emph{2023 IEEE International Conference on Robotics and Automation (ICRA)}, pages 9493--9500. IEEE, 2023.

\bibitem[Singh et~al.(2023)Singh, Blukis, Mousavian, Goyal, Xu, Tremblay, Fox, Thomason, and Garg]{singh2023progprompt}
I.~Singh, V.~Blukis, A.~Mousavian, A.~Goyal, D.~Xu, J.~Tremblay, D.~Fox, J.~Thomason, and A.~Garg.
\newblock Progprompt: Generating situated robot task plans using large language models.
\newblock In \emph{2023 IEEE International Conference on Robotics and Automation (ICRA)}, pages 11523--11530. IEEE, 2023.

\bibitem[Yu et~al.(2023)Yu, Gileadi, Fu, Kirmani, Lee, Arenas, Chiang, Erez, Hasenclever, Humplik, et~al.]{yu2023language}
W.~Yu, N.~Gileadi, C.~Fu, S.~Kirmani, K.-H. Lee, M.~G. Arenas, H.-T.~L. Chiang, T.~Erez, L.~Hasenclever, J.~Humplik, et~al.
\newblock Language to rewards for robotic skill synthesis.
\newblock \emph{arXiv preprint arXiv:2306.08647}, 2023.

\bibitem[Sha et~al.(2023)Sha, Mu, Jiang, Chen, Xu, Luo, Li, Tomizuka, Zhan, and Ding]{sha2023languagempc}
H.~Sha, Y.~Mu, Y.~Jiang, L.~Chen, C.~Xu, P.~Luo, S.~E. Li, M.~Tomizuka, W.~Zhan, and M.~Ding.
\newblock Languagempc: Large language models as decision makers for autonomous driving.
\newblock \emph{arXiv preprint arXiv:2310.03026}, 2023.

\bibitem[Mirchandani et~al.(2023)Mirchandani, Xia, Florence, Ichter, Driess, Arenas, Rao, Sadigh, and Zeng]{mirchandani2023large}
S.~Mirchandani, F.~Xia, P.~Florence, B.~Ichter, D.~Driess, M.~G. Arenas, K.~Rao, D.~Sadigh, and A.~Zeng.
\newblock Large language models as general pattern machines.
\newblock \emph{arXiv preprint arXiv:2307.04721}, 2023.

\bibitem[Arenas et~al.(2023)Arenas, Xiao, Singh, Jain, Ren, Vuong, Varley, Herzog, Leal, Kirmani, et~al.]{arenas2023prompt}
M.~G. Arenas, T.~Xiao, S.~Singh, V.~Jain, A.~Z. Ren, Q.~Vuong, J.~Varley, A.~Herzog, I.~Leal, S.~Kirmani, et~al.
\newblock How to prompt your robot: A promptbook for manipulation skills with code as policies.
\newblock In \emph{Towards Generalist Robots: Learning Paradigms for Scalable Skill Acquisition@ CoRL2023}, 2023.

\bibitem[Zha et~al.(2023)Zha, Cui, Lin, Kwon, Arenas, Zeng, Xia, and Sadigh]{zha2023distilling}
L.~Zha, Y.~Cui, L.-H. Lin, M.~Kwon, M.~G. Arenas, A.~Zeng, F.~Xia, and D.~Sadigh.
\newblock Distilling and retrieving generalizable knowledge for robot manipulation via language corrections.
\newblock \emph{arXiv preprint arXiv:2311.10678}, 2023.

\bibitem[Liang et~al.(2024)Liang, Xia, Yu, Zeng, Arenas, Attarian, Bauza, Bennice, Bewley, Dostmohamed, et~al.]{liang2024learning}
J.~Liang, F.~Xia, W.~Yu, A.~Zeng, M.~G. Arenas, M.~Attarian, M.~Bauza, M.~Bennice, A.~Bewley, A.~Dostmohamed, et~al.
\newblock Learning to learn faster from human feedback with language model predictive control.
\newblock \emph{arXiv preprint arXiv:2402.11450}, 2024.

\bibitem[Huang et~al.(2023)Huang, Wang, Zhang, Li, Wu, and Fei-Fei]{huang2023voxposer}
W.~Huang, C.~Wang, R.~Zhang, Y.~Li, J.~Wu, and L.~Fei-Fei.
\newblock Voxposer: Composable 3d value maps for robotic manipulation with language models.
\newblock \emph{arXiv preprint arXiv:2307.05973}, 2023.

\bibitem[Nasiriany et~al.(2024)Nasiriany, Xia, Yu, Xiao, Liang, Dasgupta, Xie, Driess, Wahid, Xu, et~al.]{nasiriany2024pivot}
S.~Nasiriany, F.~Xia, W.~Yu, T.~Xiao, J.~Liang, I.~Dasgupta, A.~Xie, D.~Driess, A.~Wahid, Z.~Xu, et~al.
\newblock Pivot: Iterative visual prompting elicits actionable knowledge for vlms.
\newblock \emph{arXiv preprint arXiv:2402.07872}, 2024.

\bibitem[Belkhale et~al.(2024)Belkhale, Ding, Xiao, Sermanet, Vuong, Tompson, Chebotar, Dwibedi, and Sadigh]{belkhale2024rt}
S.~Belkhale, T.~Ding, T.~Xiao, P.~Sermanet, Q.~Vuong, J.~Tompson, Y.~Chebotar, D.~Dwibedi, and D.~Sadigh.
\newblock Rt-h: Action hierarchies using language.
\newblock \emph{arXiv preprint arXiv:2403.01823}, 2024.

\bibitem[Shah et~al.(2023{\natexlab{a}})Shah, Equi, Osi{\'n}ski, Xia, Ichter, and Levine]{shah2023navigation}
D.~Shah, M.~R. Equi, B.~Osi{\'n}ski, F.~Xia, B.~Ichter, and S.~Levine.
\newblock Navigation with large language models: Semantic guesswork as a heuristic for planning.
\newblock In \emph{Conference on Robot Learning}, pages 2683--2699. PMLR, 2023{\natexlab{a}}.

\bibitem[Shah et~al.(2023{\natexlab{b}})Shah, Osi{\'n}ski, Levine, et~al.]{shah2023lm}
D.~Shah, B.~Osi{\'n}ski, S.~Levine, et~al.
\newblock Lm-nav: Robotic navigation with large pre-trained models of language, vision, and action.
\newblock In \emph{Conference on robot learning}, pages 492--504. PMLR, 2023{\natexlab{b}}.

\bibitem[Tang et~al.(2023)Tang, Yu, Tan, Zen, Faust, and Harada]{tang2023saytap}
Y.~Tang, W.~Yu, J.~Tan, H.~Zen, A.~Faust, and T.~Harada.
\newblock Saytap: Language to quadrupedal locomotion.
\newblock \emph{arXiv preprint arXiv:2306.07580}, 2023.

\bibitem[Ouyang et~al.(2024)Ouyang, Li, Li, Li, Yu, Sreenath, and Wu]{ouyang2024long}
Y.~Ouyang, J.~Li, Y.~Li, Z.~Li, C.~Yu, K.~Sreenath, and Y.~Wu.
\newblock Long-horizon locomotion and manipulation on a quadrupedal robot with large language models.
\newblock \emph{arXiv preprint arXiv:2404.05291}, 2024.

\bibitem[Chen et~al.(2022)Chen, Sharma, Levine, and Finn]{chen2022you}
A.~Chen, A.~Sharma, S.~Levine, and C.~Finn.
\newblock You only live once: Single-life reinforcement learning.
\newblock \emph{Advances in Neural Information Processing Systems}, 35:\penalty0 14784--14797, 2022.

\bibitem[Zeng et~al.(2022)Zeng, Attarian, Ichter, Choromanski, Wong, Welker, Tombari, Purohit, Ryoo, Sindhwani, et~al.]{zeng2022socratic}
A.~Zeng, M.~Attarian, B.~Ichter, K.~Choromanski, A.~Wong, S.~Welker, F.~Tombari, A.~Purohit, M.~Ryoo, V.~Sindhwani, et~al.
\newblock Socratic models: Composing zero-shot multimodal reasoning with language.
\newblock \emph{arXiv preprint arXiv:2204.00598}, 2022.

\bibitem[Min et~al.(2022)Min, Lyu, Holtzman, Artetxe, Lewis, Hajishirzi, and Zettlemoyer]{min2022rethinking}
S.~Min, X.~Lyu, A.~Holtzman, M.~Artetxe, M.~Lewis, H.~Hajishirzi, and L.~Zettlemoyer.
\newblock Rethinking the role of demonstrations: What makes in-context learning work?
\newblock \emph{arXiv preprint arXiv:2202.12837}, 2022.

\bibitem[Dong et~al.(2022)Dong, Li, Dai, Zheng, Wu, Chang, Sun, Xu, and Sui]{dong2022survey}
Q.~Dong, L.~Li, D.~Dai, C.~Zheng, Z.~Wu, B.~Chang, X.~Sun, J.~Xu, and Z.~Sui.
\newblock A survey on in-context learning.
\newblock \emph{arXiv preprint arXiv:2301.00234}, 2022.

\end{thebibliography}

\newpage
\appendix
\section{Appendix}
\label{sec:app}

\subsection{Skill Details and Hyperparameters}
\label{sec:app-skills}

We obtain different behaviors from the default controller by modulating the parameters passed in. Specifically, we control x- and y-velocity in the robot frame, gait type, body height, yaw speed, and duration to achieve different skills. The parameters used for each skill are described in the tables below, along with the duration corresponding to each magnitude. After the action is done executing, the robot will stay frozen in the position it was left in at the end of the last action, e.g. if the last action was to crawl, the robot will stay low to the ground. We additionally provide our GPT-4o query hyperparameters.

\begin{table}[!htbp]
\centering
\caption{Behavior Parameters}
\begin{tabular}{|c|c|c|c|c|c|c|}
\hline
\textbf{Parameter} & \textbf{Walk} & \textbf{Climb} & \textbf{Crawl} & \textbf{Left Turn} & \textbf{Right Turn} & \textbf{Backward} \\
\hline
X-Velocity (Small) & 0.4 m/s & 0.6 m/s & 0.3 m/s & 0 m/s & 0 m/s & -0.3 m/s \\
X-Velocity (Medium) & 0.6 m/s & 0.6 m/s & 0.3 m/s & 0 m/s & 0 m/s & -0.3 m/s \\
X-Velocity (Large) & 0.6 m/s & 0.6 m/s & 0.3 m/s & 0 m/s & 0 m/s & -0.3 m/s \\
Y-Velocity & 0 m/s & 0 m/s & 0 m/s & 0 m/s & 0 m/s & 0 m/s \\
Gait Type & 1 & 3 & 1 & 1 & 1 & 1 \\
Body Height & 0 m & 0 m & -0.3 m & 0 m & 0 m & 0 m \\
Yaw Speed & 0 rad/s & 0 rad/s & 0 rad/s & 0.3 rad/s & -0.3 rad/s & 0 rad/s \\
Duration (Small) & 3 s & 6 s & 2 s & 2.5 s & 2.5 s & 1.5 s \\
Duration (Medium) & 5 s & 9 s & 3 s & 3.5 s & 3.5 s & 2.5 s \\
Duration (Large) & 7 s & 12 s & 4 s & 4.5 s & 4.5 s & 5 s \\
\hline
\end{tabular}
\end{table}

\begin{table}[!htbp]
    \centering
    \caption{GPT-4o Query Hyperparameters}
    \begin{tabular}{|l|l|}
        \hline
        \multicolumn{1}{|c|}{Parameter} & \multicolumn{1}{c|}{Value} \\
        \hline
        Temperature & 0.7 \\
        Top P & 0.95 \\
        Max Tokens & 800 \\
        \hline
    \end{tabular}
\end{table}

\subsection{Prompts and Logs}
\label{sec:app-logs}

In the following, we include the prompts used for \ours, where the text highlighted in grey indicates text that is used for all comparison methods (including No History and No Multi-Step Plan). The text in green corresponds to the prompting for reasoning over history and is included in \ours and the No Multi-Step Plan prompts. The text highlighted in blue corresponds to prompting for multi-step planning, which is included in \ours and No History. The text in yellow corresponds to reasoning over the historical multi-step plan and is included only in the full \ours prompt. When included, the ICL prompt went immediately after the first paragraph of the initial prompt and consisted of a short explanation followed by example egocentric views and one or two actions the robot might take when it each view. For \ours and No Multi-Step Plan methods, the ``Initial Prompt'' below was given at the start and repeated after every six responses. Otherwise the ``Successive Prompt'' was given in all queries after the first. Note that we used GPT-4o as the VLM for all of our experiments, and additional prompt tuning may be necessary for other VLMs. Anecdotally, we tried using the Gemini Flash model and found that it did not reason as effectively with these prompts.

In the pages afterward, we display a full example log with \ours on the Outdoor 2 obstacle course, where the prompts and input images are included in blue and the output of the VLM at each step is included in green. See our anonymous website for videos of our results.



\section*{Supplementary material}

**Initial Prompt:**

You are controlling a robot dog to get through obstacles in this scene. The robot dog is the size of a golden retriever. The goal is to make forward progress in this starting direction and arrive at the location of a red chew toy on the other side of the obstacles in front of you. After each action you command the robot to take, you will be given a new image taken from the head of the robot dog. Each time you receive an image, you must command one of the following six actions: Walk, Crawl, Left, Right, Backward, Climb. You should command Crawl when it should go under an obstacle, command Climb when it should go over the obstacle. Examples of obstacles the robot can safely Climb over include stairs, small suitcases, cushions, small logs, and small boxes. Do not try to Crawl under these. Examples of obstacles the robot can safely Crawl under include furniture (if there is sufficient space) and suspended tree branches. Do not try to Climb furniture and suspended tree branches. Command Left when it should go left to get around an obstacle, and Right when it should go right to get around the obstacle. Right/Left command inplace turns, meaning that the robot will change the direction it is facing but not its position. This means that if you command a turn, you need to move forwards in order to move in the new direction. Do not try to turn Left or Right when you confidently believe you are underneath or above an obstacle. You should command the robot Backward if the camera is obscured and it is not clear what other action to take. If there is space to walk forward in front of you, you should command Walk. If you ever command Backward, do not try to do the same action you did before you commanded Backward again immediately (you may do so later once you are in a different scenario). If it is possible to go under the obstacle, try to Crawl, and if it is possible to go over an obstacle, try to Climb, but only if it seems like a clear path after the obstacle. If not, it might be a dead end and it might be better to move Backward. Keep in mind that the camera is on the head of the robot, and the whole body of the robot is 3 feet behind and also needs to clear the obstacle. Even if the camera view is clear, the body of the robot might be stuck: in these cases, consider Backward or repeating the same action as before. The task can be completed with the right sequence of actions, so if the robot is not succeeding, you should use the history of actions and experiences to figure out why and command different actions. When you chose Walk, Crawl, or Climb, you will move directly forward, so you may want to slightly turn to make sure you will move in the right direction before doing so.
    In your first response, describe a multi-step plan using the available actions specified above for completing the task and give reasons that plan might fail. At each subsequent step, the beginning of your response should use the state of the robot, including its current position and orientation, to reason through what progress the robot has made with the plan. You should reconsider your multi-step plan of actions to take, keeping in mind the history of what actions you have selected and the progress in the scene. If the plan seems to be working, keep following actions from that plan. If an action from the plan does not seem to be working, either revise the plan or try a making a new multi-step plan with different actions. For example, if you have had many tries of Climbing over an obstacle and the robot is unable to make reliable forward progress, you might want to try going Backward or Crawling or going Left or Right around the obstacle.
    Each time you give a response, the third to last word of your response should be a yes or no indicating whether progress was made on the prior state with the previous command. The second to last word of your output should be the action to take. The last word of your response should be a magnitude

(Small, Medium, Large) that specifies how much of that action to take. For Right or Left commands, magnitude Small corresponds to turning that direction a full camera view amount. For other commands, magnitude Small corresponds to taking that action for around one second. Once again, the last three* words of your response must be in the format yes/no Walk/Crawl/Left/Right/Backward/Climb Small/Medium/Large.

**Successive Prompt:**

Based on your last action, the robot has moved. Here is what the robot currently sees. If the view is exactly the same as before, this is likely because your last action was not appropriate for this situation and the body of the robot is probably stuck (maybe because of an unobserved obstacle). Keep in mind that this is the view from the head of the robot. Even if the camera view is clear, the body of the robot might be stuck. In these cases, consider Backward. Remember your ultimate objective of making progress in the direction the robot initially started in so that you eventually arrive at the red chew toy. Use the state of the robot including its current position and orientation to reason through what progress the robot has made with the plan. You should reconsider your multi-step plan of actions to take, keeping in mind the history of what actions you have selected and the progress in the scene. If the plan** is working well, keep following actions from that plan, making slight adjustments if necessary. If the plan** is not working, try a new multi-step plan with different actions. For example, if you had many tries of Climbing over an obstacle and the robot is unable to make reliable forward progress, you might want to try going Backward or Crawling or going Left or Right around the obstacle. End your response with the first action in the chosen plan and a magnitude in the format specified earlier.

---

*For No History: "two"
**For No Multi-Step Plan: "your old sequence of actions"

You are controlling a robot dog to get through obstacles in this scene. The robot dog is the size of a golden retriever. The goal is to make forward progress in this starting direction and arrive at the location of a red chew toy on the other side of the obstacles in front of you. After each action you command the robot to take, you will be given a new image taken from the head of the robot dog. Each time you receive an image, you must command one of the following six actions: Walk, Crawl, Left, Right, Backward, Climb. You should command Crawl when it should go under an obstacle, command Climb when it should go over the obstacle. Examples of obstacles the robot can safely Climb over include stairs, small suitcases, cushions, small logs, and small boxes. Do not try to Crawl under these. Examples of obstacles the robot can safely Crawl under include furniture (if there is sufficient space) and suspended tree branches. Do not try to Climb furniture and suspended tree branches. Command Left when it should go left to get around an obstacle, and Right when it should go right to get around the obstacle. Right/Left command inplace turns, meaning that the robot will change the direction it is facing but not its position. This means that if you command a turn, you need to move forwards in order to move in the new direction. Do not try to turn Left or Right when you confidently believe you are underneath or above an obstacle. You should command the robot Backward if the camera is obscured and it is not clear what other action to take. If there is space to walk forward in front of you, you should command Walk. If you ever command Backward, do not try to do the same action you did before you commanded Backward again immediately (you may do so later once you are in a different scenario). If it is possible to go under the obstacle, try to Crawl, and if it is possible to go over an obstacle, try to Climb, but only if it seems like a clear path after the obstacle. If not, it might be a dead end and it might be better to move Backward. Keep in mind that the camera is on the head of the robot, and the whole body of the robot is 3 feet behind and also needs to clear the obstacle. Even if the camera view is clear, the body of the robot might be stuck: in these cases, consider Backward or repeating the same action as before. The task can be completed with the right sequence of actions, so if the robot is not succeeding, you should use the history of actions and experiences to figure out why and command different actions. When you chose Walk, Crawl, or Climb, you will move directly forward, so you may want to slightly turn to make sure you will move in the right direction before doing so. In your first response, describe a multi-step plan using the available actions specified above for completing the task and give reasons that plan might fail. At each subsequent step, the beginning of your response should use the state of the robot, including its current position and orientation, to reason through what progress the robot has made with the plan. You should reconsider your multi-step plan of actions to take, keeping in mind the history of what actions you have selected and the progress in the scene. If the plan seems to be working, keep following actions from that plan. If an action from the plan does not seem to be working, either revise the plan or try a making a new multi-step plan with different actions. For example, if you have had many tries of Climbing over an obstacle and the robot is unable to make reliable forward progress, you might want to try going Backward or Crawling or going Left or Right around the obstacle. Each time you give a response, the third to last word of your response should be a yes or no indicating whether progress was made on the prior state with the previous command. The second to last word of your output should be the action to take. The last word of your response should be a magnitude (Small, Medium, Large) that specifies how much of that action to take. For Right or Left commands, magnitude Small corresponds to turning that direction a full camera view amount. For other commands, magnitude Small corresponds to taking that action for around one second. Once again, the last three words of your response must be in the format yes/no Walk/Crawl/Left/Right/Backward/Climb Small/Medium/Large.

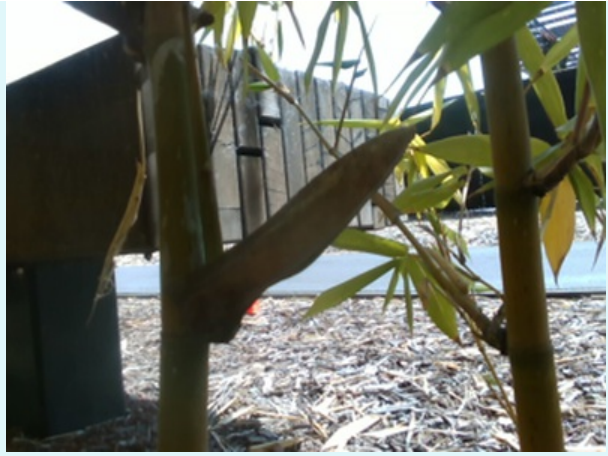

**Multi-Step Plan:**

1. **Crawl (Small)**: The robot dog should crawl under the bamboo stems in front of it. This will allow it to make progress without getting stuck or needing to navigate around the bamboo.
2. **Walk (Medium)**: After clearing the bamboo, the robot dog should walk forward to cover more ground towards the red chew toy.
3. **Left (Small)**: If there are obstacles directly ahead, the robot dog should turn left to see if there is a clearer path.
4. **Walk (Medium)**: Continue walking forward if the path is clear.
5. **Right (Small)**: If there are still obstacles, the robot dog should turn right to check for another clear path.

**Reasons the Plan Might Fail:**

- The bamboo stems might be too low or too close together for the robot dog to crawl under.
- There may be additional obstacles not visible in the initial image that block the robot dog's path.
- The robot dog might get stuck on uneven ground or hidden objects while crawling or walking.

**Current Action:**

Let's begin with the first step of the plan.

yes Crawl Small

Based on your last action, the robot has moved. Here is what the robot currently sees. If the view is exactly the same as before, this is likely because your last action was not appropriate for this situation and the body of the robot is probably stuck (maybe because of an unobserved obstacle). Keep in mind that this is the view from the head of the robot. Even if the camera view is clear, the body of the robot might be stuck. In these cases, consider Backward. Remember your ultimate objective of making progress in the direction the robot initially started in so that you eventually arrive at the red chew toy. Use the state of the robot including its current position and orientation to reason through what progress the robot has made with the plan. You reconsider your new multi-step plan of actions to take, keeping in mind the history of what actions you have selected and the progress in the scene. If the plan is working well, keep following actions from that plan, making slight adjustments if necessary. If the plan is not working, try a new multi-step plan with different actions. For example, if you had many tries of Climbing over an obstacle and the robot is unable to make reliable forward progress, you might want to try going Backward or Crawling or going Left or Right around the obstacle. End your response with the first action in the chosen plan and a magnitude in the format specified earlier.

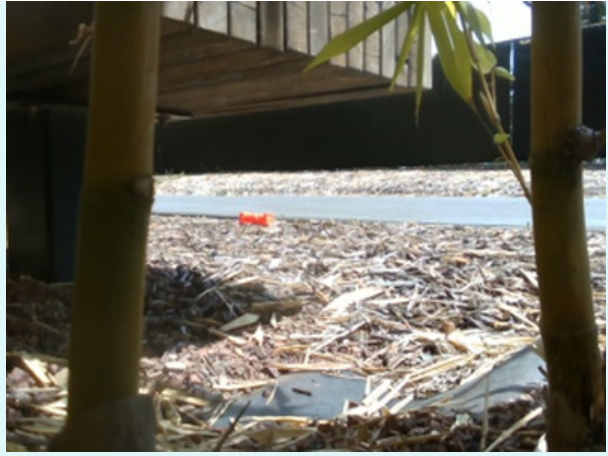

## Analysis:

The robot has successfully crawled under the bamboo stems, as evidenced by the new view. The red chew toy is now visible in the distance, indicating forward progress. However, there is still an obstacle in the form of a low platform or bench ahead. The space underneath this platform appears large enough for the robot to crawl under.

## Revised Multi-Step Plan:

1. **Crawl (Small)**: Continue crawling forward to move under the low platform or bench.
2. **Walk (Medium)**: Once clear of the platform, the robot should walk forward to close the remaining distance to the chew toy.
3. **Left (Small)**: If there are obstacles directly ahead, turn left to find a clear path.
4. **Walk (Medium)**: Continue walking forward if the path is clear.
5. **Right (Small)**: If there are still obstacles, turn right to check for another clear path.

## Next Action:

Let's proceed with the first step of the revised plan.

yes Crawl Small

Based on your last action, the robot has moved. Here is what the robot currently sees. If the view is exactly the same as before, this is likely because your last action was not appropriate for this situation and the body of the robot is probably stuck (maybe because of an unobserved obstacle). Keep in mind that this is the view from the head of the robot. Even if the camera view is clear, the body of the robot might be stuck. In these cases, consider Backward. Remember your ultimate objective of making progress in the direction the robot initially started in so that you eventually arrive at the red chew toy. Use the state of the robot including its current position and orientation to reason through what progress the robot has made with the plan. You reconsider your new multi-step plan of actions to take, keeping in mind the history of what actions you have selected and the progress in the scene. If the plan is working well, keep following actions from that plan, making slight adjustments if necessary. If the plan is not working, try a new multi-step plan with different actions. For example, if you had many tries of Climbing over an obstacle and the robot is unable to make reliable forward progress, you might want to try going Backward or Crawling or going Left or Right around the obstacle. End your response with the first action in the chosen plan and a magnitude in the format specified earlier.

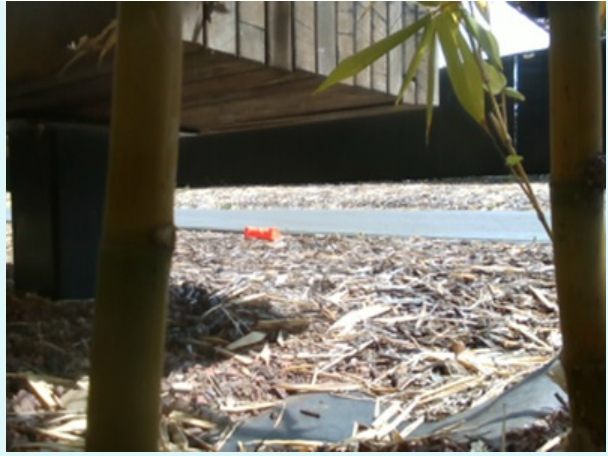

## Analysis:

The view has not changed, indicating that the robot might be stuck. The body of the robot may not have cleared the bamboo stems completely or there could be another unseen obstacle behind or beneath the platform.

## Revised Multi-Step Plan:

1. **Backward (Small)**: Move backward to ensure the robot is completely clear of any obstacles it might be stuck on.
2. **Crawl (Medium)**: After moving backward, attempt to crawl under the platform again to ensure the robot can clear the obstacle.

## Next Action:

Let's proceed with the first step of the revised plan.

no Backward Small

Based on your last action, the robot has moved. Here is what the robot currently sees. If the view is exactly the same as before, this is likely because your last action was not appropriate for this situation and the body of the robot is probably stuck (maybe because of an unobserved obstacle). Keep in mind that this is the view from the head of the robot. Even if the camera view is clear, the body of the robot might be stuck. In these cases, consider Backward. Remember your ultimate objective of making progress in the direction the robot initially started in so that you eventually arrive at the red chew toy. Use the state of the robot including its current position and orientation to reason through what progress the robot has made with the plan. You reconsider your new multi-step plan of actions to take, keeping in mind the history of what actions you have selected and the progress in the scene. If the plan is working well, keep following actions from that plan, making slight adjustments if necessary. If the plan is not working, try a new multi-step plan with different actions. For example, if you had many tries of Climbing over an obstacle and the robot is unable to make reliable forward progress, you might want to try going Backward or Crawling or going Left or Right around the obstacle. End your response with the first action in the chosen plan and a magnitude in the format specified earlier.

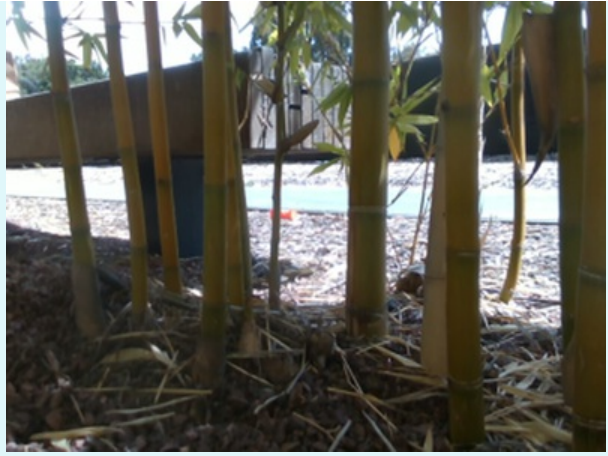

## Analysis:

The robot has successfully moved backward, revealing more of the bamboo stems and the low platform. The red chew toy is still visible in the distance. The robot is now in a better position to reattempt crawling under the platform.

## Revised Multi-Step Plan:

1. **Left (Small)**: Turn left to better align the robot with a clearer path through the bamboo stems.
2. **Crawl (Medium)**: After turning, crawl forward to move under the low platform.
3. **Walk (Medium)**: Once clear of the platform, walk forward to close the remaining distance to the chew toy.
4. **Left (Small)**: If there are obstacles directly ahead, turn left to find a clear path.
5. **Walk (Medium)**: Continue walking forward if the path is clear.

## Next Action:

Let's proceed with the first step of the revised plan.

yes Left Small

Based on your last action, the robot has moved. Here is what the robot currently sees. If the view is exactly the same as before, this is likely because your last action was not appropriate for this situation and the body of the robot is probably stuck (maybe because of an unobserved obstacle). Keep in mind that this is the view from the head of the robot. Even if the camera view is clear, the body of the robot might be stuck. In these cases, consider Backward. Remember your ultimate objective of making progress in the direction the robot initially started in so that you eventually arrive at the red chew toy. Use the state of the robot including its current position and orientation to reason through what progress the robot has made with the plan. You reconsider your new multi-step plan of actions to take, keeping in mind the history of what actions you have selected and the progress in the scene. If the plan is working well, keep following actions from that plan, making slight adjustments if necessary. If the plan is not working, try a new multi-step plan with different actions. For example, if you had many tries of Climbing over an obstacle and the robot is unable to make reliable forward progress, you might want to try going Backward or Crawling or going Left or Right around the obstacle. End your response with the first action in the chosen plan and a magnitude in the format specified earlier.

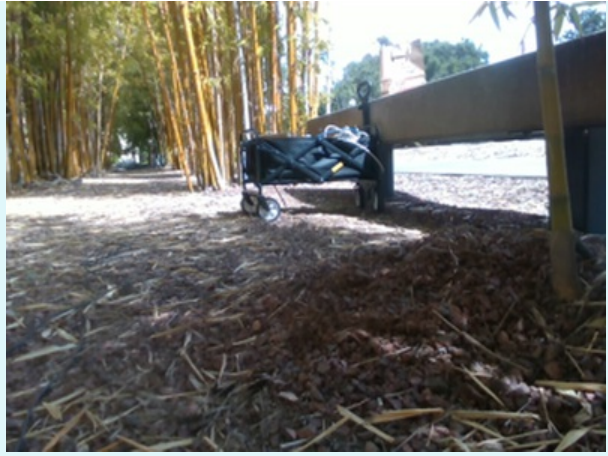

### Analysis:

The robot has successfully turned left, revealing a clearer path ahead. There is a new obstacle, a cart, which the robot can likely climb over. The red chew toy is still visible in the distance, indicating the direction to progress.

### Revised Multi-Step Plan:

1. **Climb (Medium)**: Climb over the cart to make progress towards the red chew toy.
2. **Walk (Medium)**: Once clear of the cart, walk forward to close the remaining distance to the chew toy.
3. **Left (Small)**: If there are obstacles directly ahead, turn left to find a clear path.
4. **Walk (Medium)**: Continue walking forward if the path is clear.
5. **Right (Small)**: If there are still obstacles, turn right to check for another clear path.

### Next Action:

Let's proceed with the first step of the revised plan.

yes Climb Medium

Based on your last action, the robot has moved. Here is what the robot currently sees. If the view is exactly the same as before, this is likely because your last action was not appropriate for this situation and the body of the robot is probably stuck (maybe because of an unobserved obstacle). Keep in mind that this is the view from the head of the robot. Even if the camera view is clear, the body of the robot might be stuck. In these cases, consider Backward. Remember your ultimate objective of making progress in the direction the robot initially started in so that you eventually arrive at the red chew toy. Use the state of the robot including its current position and orientation to reason through what progress the robot has made with the plan. You reconsider your new multi-step plan of actions to take, keeping in mind the history of what actions you have selected and the progress in the scene. If the plan is working well, keep following actions from that plan, making slight adjustments if necessary. If the plan is not working, try a new multi-step plan with different actions. For example, if you had many tries of Climbing over an obstacle and the robot is unable to make reliable forward progress, you might want to try going Backward or Crawling or going Left or Right around the obstacle. End your response with the first action in the chosen plan and a magnitude in the format specified earlier.

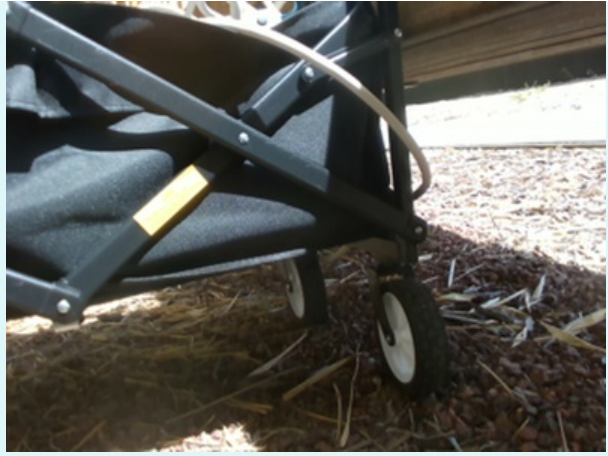

**Analysis:**

The robot has moved forward and is now partially under the cart, indicating that the climb was not entirely successful. The body of the robot may be stuck under the cart.

**Revised Multi-Step Plan:**

1. **Backward (Small)**: Move backward to clear the robot from being stuck under the cart.
2. **Right (Small)**: Turn right to align with a path around the cart.
3. **Walk (Medium)**: Walk forward to bypass the cart and continue towards the red chew toy.
4. **Left (Small)**: If there are obstacles directly ahead, turn left to find a clear path.
5. **Walk (Medium)**: Continue walking forward if the path is clear.

**Next Action:**

Let's proceed with the first step of the revised plan.

no Backward Small

As a reminder, you are controlling a robot dog to get through obstacles in this scene. The robot dog is the size of a golden retriever. The goal is to make forward progress in this starting direction and arrive at the location of a red chew toy on the other side of the obstacles in front of you. After each action you command the robot to take, you will be given a new image taken from the head of the robot dog. Each time you receive an image, you must command one of the following six actions: Walk, Crawl, Left, Right, Backward, Climb. You should command Crawl when it should go under an obstacle, command Climb when it should go over the obstacle. Examples of obstacles the robot can safely Climb over include stairs, small suitcases, cushions, small logs, and small boxes. Do not try to Crawl under these. Examples of obstacles the robot can safely Crawl under include furniture (if there is sufficient space) and suspended tree branches. Do not try to Climb furniture and suspended tree branches. Command Left when it should go left to get around an obstacle, and Right when it should go right to get around the obstacle. Right/Left command inplace turns, meaning that the robot will change the direction it is facing but not its position. This means that if you command a turn, you need to move forwards in order to move in the new direction. Do not try to turn Left or Right when you confidently believe you are underneath or above an obstacle. You should command the robot Backward if the camera is obscured and it is not clear what other action to take. If there is space to walk forward in front of you, you should command Walk. If you ever command Backward, do not try to do the same action you did before you commanded Backward again immediately (you may do so later once you are in a different scenario). If it is possible to go under the obstacle, try to Crawl, and if it is possible to go over an obstacle, try to Climb, but only if it seems like a clear path after the obstacle. If not, it might be a dead end and it might be better to move Backward.

Keep in mind that the camera is on the head of the robot, and the whole body of the robot is 3 feet behind and also needs to clear the obstacle. Even if the camera view is clear, the body of the robot might be stuck: in these cases, consider Backward or repeating the same action as before. The task can be completed with the right sequence of actions, so if the robot is not succeeding, you should use the history of actions and experiences to figure out why and command different actions. When you chose Walk, Crawl, or Climb, you will move directly forward, so you may want to slightly turn to make sure you will move in the right direction before doing so. In your first response, describe a multi-step plan using the available actions specified above for completing the task and give reasons that plan might fail. At each subsequent step, the beginning of your response should use the state of the robot, including its current position and orientation, to reason through what progress the robot has made with the plan. You should reconsider your multi-step plan of actions to take, keeping in mind the history of what actions you have selected and the progress in the scene. If the plan seems to be working, keep following actions from that plan. If an action from the plan does not seem to be working, either revise the plan or try a making a new multi-step plan with different actions. For example, if you have had many tries of Climbing over an obstacle and the robot is unable to make reliable forward progress, you might want to try going Backward or Crawling or going Left or Right around the obstacle. Each time you give a response, the third to last word of your response should be a yes or no indicating whether progress was made on the prior state with the previous command. The second to last word of your output should be the action to take. The last word of your response should be a magnitude (Small, Medium, Large) that specifies how much of that action to take. For Right or Left commands, magnitude Small corresponds to turning that direction a full camera view amount. For other commands, magnitude Small corresponds to taking that action for around one second. Once again, the last three words of your response must be in the format yes/no Walk/Crawl/Left/Right/Backward/Climb Small/Medium/Large.

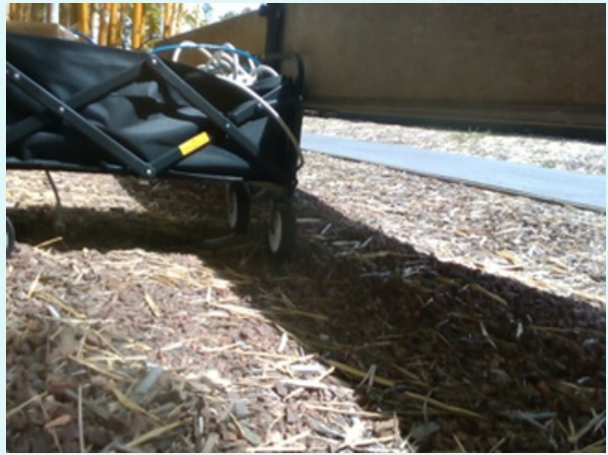

### Analysis:

The robot has successfully moved backward and now has a clearer view of the cart and the path around it. The path to the right seems to provide more space to navigate around the cart.

### Revised Multi-Step Plan:

1. **Right (Small)**: Turn right to align with the clearer path around the cart.
2. **Walk (Medium)**: Walk forward to bypass the cart and continue towards the red chew toy.
3. **Left (Small)**: If there are obstacles directly ahead, turn left to find a clear path.
4. **Walk (Medium)**: Continue walking forward if the path is clear.
5. **Right (Small)**: If there are still obstacles, turn right to check for another clear path.

### Next Action:

Let's proceed with the first step of the revised plan.

yes Right Small

Based on your last action, the robot has moved. Here is what the robot currently sees. If the view is exactly the same as before, this is likely because your last action was not appropriate for this situation and the body of the robot is probably stuck (maybe because of an unobserved obstacle). Keep in mind that this is the view from the head of the robot. Even if the camera view is clear, the body of the robot might be stuck. In these cases, consider Backward. Remember your ultimate objective of making progress in the direction the robot initially started in so that you eventually arrive at the red chew toy. Use the state of the robot including its current position and orientation to reason through what progress the robot has made with the plan. You reconsider your new multi-step plan of actions to take, keeping in mind the history of what actions you have selected and the progress in the scene. If the plan is working well, keep following actions from that plan, making slight adjustments if necessary. If the plan is not working, try a new multi-step plan with different actions. For example, if you had many tries of Climbing over an obstacle and the robot is unable to make reliable forward progress, you might want to try going Backward or Crawling or going Left or Right around the obstacle. End your response with the first action in the chosen plan and a magnitude in the format specified earlier.

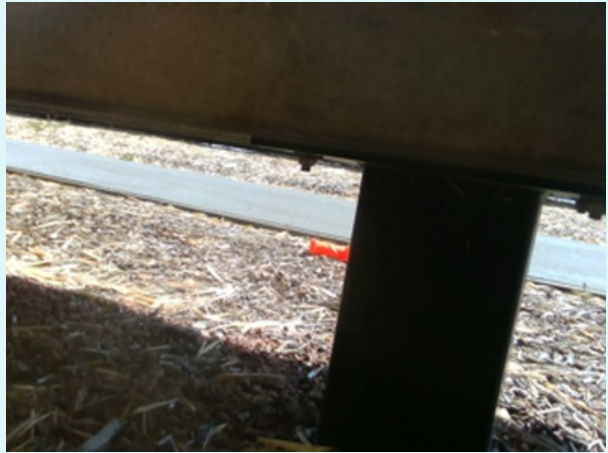

### Analysis:

The robot has successfully turned right, providing a clearer view under the platform. The red chew toy is still visible in the distance. The space under the platform appears to be large enough for the robot to crawl under it.

### Revised Multi-Step Plan:

1. **Crawl (Medium)**: Crawl forward to move under the platform.
2. **Walk (Medium)**: Once clear of the platform, walk forward to close the remaining distance to the chew toy.
3. **Left (Small)**: If there are obstacles directly ahead, turn left to find a clear path.
4. **Walk (Medium)**: Continue walking forward if the path is clear.
5. **Right (Small)**: If there are still obstacles, turn right to check for another clear path.

### Next Action:

Let's proceed with the first step of the revised plan.

yes Crawl Medium

Based on your last action, the robot has moved. Here is what the robot currently sees. If the view is exactly the same as before, this is likely because your last action was not appropriate for this situation and the body of the robot is probably

stuck (maybe because of an unobserved obstacle). Keep in mind that this is the view from the head of the robot. Even if the camera view is clear, the body of the robot might be stuck. In these cases, consider Backward. Remember your ultimate objective of making progress in the direction the robot initially started in so that you eventually arrive at the red chew toy. Use the state of the robot including its current position and orientation to reason through what progress the robot has made with the plan. You reconsider your new multi-step plan of actions to take, keeping in mind the history of what actions you have selected and the progress in the scene. If the plan is working well, keep following actions from that plan, making slight adjustments if necessary. If the plan is not working, try a new multi-step plan with different actions. For example, if you had many tries of Climbing over an obstacle and the robot is unable to make reliable forward progress, you might want to try going Backward or Crawling or going Left or Right around the obstacle. End your response with the first action in the chosen plan and a magnitude in the format specified earlier.

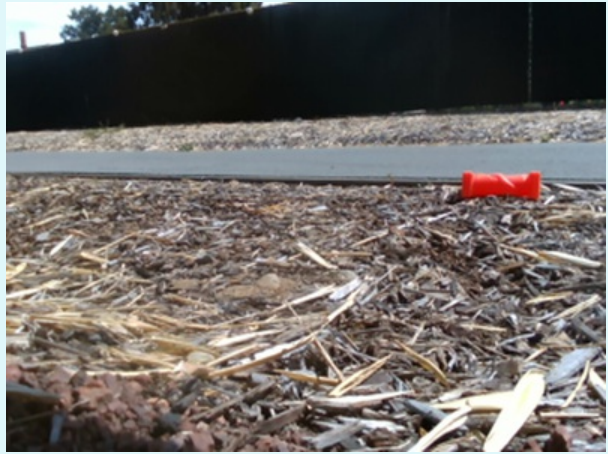

## Analysis:

The robot has successfully crawled under the platform, and the view is now clear with the red chew toy very close ahead. There are no visible obstacles between the robot and the toy.

## Revised Multi-Step Plan:

1. **Walk (Medium)**: Walk forward to reach the red chew toy.

## Next Action:

Let's proceed with the first step of the revised plan.

yes Walk Medium



\end{document}